\documentclass[11pt,letterpaper]{article}

\usepackage[T1]{fontenc}
\usepackage{lmodern}
\usepackage[margin=1in]{geometry}
\usepackage[round]{natbib}
\usepackage{amsmath,amssymb,amsthm}
\usepackage{graphicx}
\usepackage{booktabs}
\usepackage{enumitem}
\usepackage[hidelinks]{hyperref}
\hypersetup{
  pdftitle={The Winner's Curse in LLM Self-Improvement Loops: Selection Noise, Lock-in, and Acceptance Rules},
  pdfauthor={Litao Hu and Yutong Tang},
  pdfsubject={Selection noise and empirical evaluation in LLM self-improvement},
  pdfkeywords={large language models, selection bias, winner's curse, prompt optimization}
}

\newtheorem{proposition}{Proposition}
\newtheorem{assumption}{Assumption}
\title{The Winner's Curse in LLM Self-Improvement Loops:\\
Selection Noise, Lock-in, and Acceptance Rules}
\author{Litao Hu\\Meta \and Yutong Tang\\Microsoft}
\date{}

\begin{document}
\maketitle

\begin{abstract}
Self-improving LLM systems propose changes to themselves and keep those that score better on a small evaluation set.
We treat this keep-if-better step as selection under measurement noise, model the correlated errors of the candidates
in a single decision, and study empirically what happens when the evaluation set is reused. In runs where Qwen models
rewrite their own instructions and every candidate is also scored on 600 held-out items, most proposals after the
first are harmful, and the model gives the size of the winner's curse of a generation's best candidate. With a prior
from a separate pilot, it matches the average overstatement of first-generation commits in native loops, though not
setting by setting. In a pre-registered study, the final selection-set score of greedy loops exceeded held-out
accuracy by 13 to 20 points with 16 selection items and by 1 to 5 points with 256. Held-out gains grew with the
selection set on TREC but not on GSM8K, and the tested acceptance rules did not beat greedy acceptance over whole runs.
Gains measured on the selection set also exceeded held-out gains when a current model refined a competent instruction,
and in the validation scores of GEPA and MIPROv2. Scoring the starting and the current instruction on 64 items never
used for selection removes the average bias of a loop's reported gain, but single estimates remain off by about 6
points. Self-improvement studies should report held-out gains with their uncertainty.

\end{abstract}

\section{Introduction}
\label{sec:intro}

A growing class of LLM systems improves itself by editing its own artifacts. Prompt optimizers rewrite the
instructions of a pipeline \citep{zhou2023ape,yang2024opro,agrawal2025gepa}, agents revise their memories, tools and
scaffolds \citep{hu2025adas,zhang2025dgm}, and LLM-guided search rewrites programs \citep{novikov2025alphaevolve}.
These systems share one step. They propose a change, score it on a finite set of tasks, and keep it if it scores
better than what it would replace. The score that decides what is kept is also what a loop's logs and an optimizer's
validation output show as progress. Careful studies report separate test sets as well, but several recent audits
suggest that the selection score should not be taken at face value. Evolved agent harnesses do not consistently beat
test-time scaling at matched budget \citep{rethinkharness2026}, greedy acceptance commits many edits that are false
or harmful \citep{pace2026}, measured self-training gains can be artifacts of the evaluation \citep{xu2026phantom},
and a 41-paper methods audit found commonly omitted validity controls \citep{audit2026}.

Statistics predicts these findings. When the best of several noisy estimates is selected, its estimate is biased
upward. The breeder's equation, the optimizer's curse and the winner's curse of launched A/B tests are instances of
this effect \citep{falconer1996introduction,smith2006optimizer,lee2018winner}, and a decision-theoretic remedy is an adoption
threshold informed by a prior over effects \citep{berman2022false}. LLM self-improvement differs from the classical
settings: the proposer is often the system being improved, the candidates are correlated rewrites of one artifact,
one small evaluation set is reused for many generations, and paired held-out effects can be estimated.

We model the keep-if-better step as selection under measurement noise, with a random-effects structure in which the
candidates of a generation share the incumbent's error (Section~\ref{sec:model}). We distinguish two quantities
throughout. The \emph{inflation} of a program is its score on the selection set minus its held-out accuracy, and the
\emph{overstatement} of a gain is the gain measured on the selection set minus the held-out gain, that is, the final
program's inflation minus the starting program's. We study the decision inside the loop: how correlated candidate
errors distort the gain of the selected candidate, how reused items change the incumbent against which later
candidates are judged, and whether corrections that help isolated decisions improve complete runs. The first is
derived under a fresh-set model; the other two are empirical. In oracle-instrumented runs, in which Qwen models
rewrite their own instructions and every candidate is also scored on a 600-item held-out set, most proposals after
the first rewrite of the seed are harmful and the winner's curse has the size the model predicts. In native loops,
the model with a prior from a separate pilot matches the average overstatement of first-generation commits, with
errors of several points in individual settings. A pre-registered study
confirms that the inflation of a loop's final program falls as the selection set grows, from 13 to 20 points with a
reused 16-item set to 1 to 5 with 256 items. Its other hypotheses were supported only in part or not at all: held-out
gains grew with the selection set on TREC but not on GSM8K, and neither a calibrated Bayes gate nor select-then-confirm
beat greedy acceptance in whole runs. The overstatement persists when Qwen3.5-4B refines a competent instruction, and
in the validation scores of GEPA and MIPROv2 it reaches 26 and 7 points with 16 validation items.

Item-level logs support within-generation shrinkage estimates; acceptance-conditioned predictions also need a prior,
and accumulated inflation requires held-out evaluation. In our reused-set experiments, the tested rules did not
reliably improve final held-out gains.

\section{Related work}
\label{sec:related}

\paragraph{Self-improving LLM systems.}
Prompt and program optimizers search over instructions and demonstrations with an LLM as the mutation operator
\citep{zhou2023ape,yang2024opro,pryzant2023protegi,guo2024evoprompt,fernando2024promptbreeder,khattab2024dspy,
opsahlong2024mipro,yuksekgonul2024textgrad,agrawal2025gepa}. Self-referential systems apply the same idea to the code
that performs the improvement \citep{zelikman2024stop,yin2024godel}, agent-design and self-modification loops to
scaffolds and tools \citep{hu2025adas,zhang2025dgm,wang2025hgm}, and LLM-guided evolutionary search to programs
\citep{romera2024funsearch,novikov2025alphaevolve}. All of these systems decide which changes to keep on the basis
of a finite evaluation, and some already guard against its noise. GEPA scores proposals on a minibatch before a full
validation pass, ProTeGi and TRIPLE treat candidate selection as a bandit problem
\citep{pryzant2023protegi,shi2024triple}, and DGM and HGM evaluate on progressively larger task subsets. We study GEPA
and MIPROv2 directly in Section~\ref{sec:gepa}.

\paragraph{Reliability of self-improvement.}
Two recent papers are closest. PACE \citep{pace2026} runs prompt self-evolution with small Qwen2.5 models, counts
the false and harmful edits that greedy acceptance commits, and replaces greedy acceptance by an anytime-valid test
that controls the false-commit probability of each decision. SIREN \citep{siren2026} studies reporting after
budgeted adaptive search over prompts or programs that reuses benchmark items. It quantifies the resulting optimism
and gives inference for the tune-then-deploy procedure under its fixed-shortlist and stabilized-selection assumptions,
using held-out splits. Our target is the decision inside the loop instead. We model how correlated
candidate errors distort the gain of the selected candidate, measure how reused items change the incumbent against
which later candidates are judged, and ask whether corrections that help isolated decisions improve complete runs, in
our own loops and in GEPA and MIPROv2. The degradation that PACE reports without real improvements appeared in our exploratory runs but did not
replicate in our pre-registered ones. Other work audits self-training against a measured null \citep{xu2026phantom}
and documents fragile or unreplicated gains of self-improving agents
\citep{rethinkharness2026,fragility2026}. A methods audit identifies omitted validity controls in published studies
\citep{audit2026}.

\paragraph{Selection under noise.}
Our model adapts standard results, including the breeder's equation \citep{lush1937animal,
falconer1996introduction}, regression to the mean and the optimizer's curse \citep{harrison1984postdecision,
smith2006optimizer}, selection-adjusted Bayesian inference \citep{dawid1994selection,efron2011tweedie} and
inference on winners \citep{andrews2024inference}. Online experimentation faces the same launch decision, with many
candidate changes, finite traffic, a winner's curse, and priors that set the adoption threshold
\citep{lee2018winner,azevedo2020ab,berman2022false}. Noisy evolution strategies \citep{hammel1994evolution,beyer2000noise,arnold2002noisy,arnold2006general} and the
drift barrier \citep{ohta1973slightly,sung2012drift} are relatives. In machine learning, algorithm configuration handles
over-tuning and incumbent re-evaluation \citep{hutter2009paramils,
lopez2016irace}, model selection has its own selection bias \citep{cawley2010overfitting}, adaptive data analysis
bounds the overfitting caused by holdout reuse \citep{dwork2015reusable,blum2015ladder}, and reward-model
over-optimization is the corresponding phenomenon in RLHF \citep{gao2023overopt}.

\section{A selection model of keep-if-better loops}
\label{sec:model}

\paragraph{The loop.}
An \emph{artifact} $c$ (an instruction, program, memory, or scaffold) conditions a fixed solver with performance
$f(c)=\mathbb{E}_{x\sim P}[r(c,x)]$, $r\in\{0,1\}$. At generation $t$ a proposer draws $K$ candidates from the
incumbent $c_t$; each is scored on a \emph{selection set} $D=(x_1,\dots,x_n)\stackrel{iid}{\sim}P$, paired with the
incumbent:
\begin{equation}
\hat\Delta_k=\frac1n\sum_{i=1}^n\big[r(c'_k,x_i)-r(c_t,x_i)\big],
\end{equation}
an estimate of $\Delta_k=f(c'_k)-f(c_t)$. An \emph{acceptance rule} commits one candidate or none; greedy
keep-if-better commits $\hat k=\arg\max_k\hat\Delta_k$ iff $\hat\Delta_{\hat k}>0$. We study single-incumbent loops and
apply the model to the program that population-based search returns in Section~\ref{sec:gepa}.

\paragraph{Measurement noise.}
Only discordant items contribute: with $q_k=\Pr[r(c'_k,x)\ne r(c_t,x)]$, $\hat\Delta_k$ is unbiased on a
\emph{fresh} $D$ with variance $v_k/n$, $v_k=q_k-\Delta_k^2$. Rewrites of one instruction change many answers
($v\approx0.03$--$0.2$ per item in our runs), so $n$ items resolve differences of order $\sqrt{v/n}$ only.

\begin{assumption}[Gaussian selection model]\label{ass:gauss}
Given the history and a fresh selection set, $\Delta_k\stackrel{iid}{\sim}N(\mu,s^2)$ and
$\hat\Delta_k=\Delta_k+c+\eta_k$ with a shared error $c\sim N(0,e_c^2)$ (all candidates face the same incumbent and
items) and idiosyncratic errors $\eta_k\stackrel{iid}{\sim}N(0,e_\eta^2)$, with $c$, all $\eta_k$ and all $\Delta_k$
mutually independent.
\end{assumption}
$\mu$ is the \emph{proposal bias} (mean true effect of a proposed change) and $s$ the \emph{proposal spread}. The
noise terms are identified from item vectors: $e^2=e_c^2+e_\eta^2=v/n$ and $e_\eta^2=v_{cc}/n$, where
$v_{cc}=\tfrac12\mathbb{E}\,\mathrm{Var}_x[r(c'_j,x)-r(c'_k,x)]$ is half the per-item variance of the difference
between two candidates. Write $\sigma_w^2=s^2+e_\eta^2$, $h_w^2=s^2/\sigma_w^2$, $h^2=s^2/(s^2+e^2)$, and
$\kappa_K=\mathbb{E}\max_{k\le K}Z_k$ for i.i.d.\ standard normals ($\kappa_4\approx1.03$, $\kappa_{16}\approx1.77$;
the approximation $\sqrt{2\ln K}$, which gives 1.67 for $K=4$, holds only for large $K$). The posterior formulas
assume $\sigma_w^2>0$.

\subsection{The winner's curse}

\begin{proposition}[Selection differential and overstatement]\label{prop:breeder}
Under Assumption~\ref{ass:gauss}, let $\bar m$ and $\bar\Delta$ be the generation means of measured and true effects
and $\hat k=\arg\max_k\hat\Delta_k$. Then
$\mathbb{E}[\Delta_{\hat k}-\bar\Delta\mid\hat\Delta]=h_w^2(\hat\Delta_{\hat k}-\bar m)$ and
$\mathbb{E}[\hat\Delta_{\hat k}-\bar m]=\sigma_w\kappa_K$, so the true selection differential is
$h_w^2\sigma_w\kappa_K$ (the breeder's equation $R=h^2S$ when $e_c=0$). The measured gain of $\hat k$ over the
incumbent overstates its true gain by
\begin{equation}\label{eq:over}
\mathbb{E}[\hat\Delta_{\hat k}-\Delta_{\hat k}]=(1-h_w^2)\,\sigma_w\kappa_K+\mathbb{E}[c].
\end{equation}
\end{proposition}
The first part is regression to the mean of the selected candidate; Equation~\eqref{eq:over} is the optimizer's
curse \citep{smith2006optimizer,harrison1984postdecision}, with $\mathbb{E}[c]=0$ on a fresh selection set. The proposition
describes the best-measured candidate before any acceptance filter. A greedy commit is that candidate given
$\hat\Delta_{\hat k}>0$, and conditioning on acceptance raises the bias when most proposals are harmful. For a single
candidate with true effect $-e$ and noise $N(0,e^2)$, the unconditional error is zero, but the mean error of an
accepted candidate is $e\,\phi(1)/(1-\Phi(1))\approx1.53e$. Section~\ref{sec:lockin} uses the acceptance-conditioned
value.

\subsection{The Bayes rule and calibrated thresholds}

\begin{proposition}[Myopic Bayes commit]\label{prop:bayes}
Under Assumption~\ref{ass:gauss} with $(\mu,s^2,e_c^2,e_\eta^2)$ known, the rule that maximizes the expected true
improvement of one generation commits the candidate with the largest posterior mean
\begin{equation}\label{eq:post}
\mathbb{E}[\Delta_k\mid\hat\Delta]=\mu+h_w^2(\hat\Delta_k-\bar m)+\frac{h_w^2\sigma_w^2}{\sigma_w^2+Ke_c^2}(\bar m-\mu)
\end{equation}
iff that posterior mean is positive. With independent noise ($e_c=0$) and $s>0$ this is a threshold on the measured gain,
$\hat\Delta_{\hat k}>\tau^\star=-\mu\,e^2/s^2$.
\end{proposition}
This is the standard Bayes adoption rule \citep{dawid1994selection,berman2022false}. It is optimal only
myopically (ignoring a commit's effect on later generations) and with known hyperparameters, and the gates we run
(Section~\ref{sec:method}) add a safeguard that it lacks, a positive measured gain. With independent noise and
$s,e>0$, greedy acceptance ($\tau=0$) is optimal only if $\mu=0$; a fixed-level test uses $\tau=z_\alpha e$ regardless
of $\mu$ and $s$. When most proposals are harmful and candidates barely differ ($s\ll e$), the right threshold is
large; Equation~\eqref{eq:post} says how to calibrate it.

\subsection{Goodhart regime and lock-in}

\begin{proposition}[Unresolvable regime, fresh selection sets]\label{prop:goodhart}
Let $s=0$ and $\mu<0$, with proposal and noise distributions that do not change across generations, and draw a
fresh selection set each generation. Greedy commits with probability
$p=\Pr[c+e_\eta\max_kZ_k>|\mu|]$ per generation, which increases with $K$ and decreases with $n$; held-out
performance drifts down by $|\mu|p$ per generation while every commit reports a positive gain.
\end{proposition}
This is regressional Goodhart \citep{manheim2018goodhart}, the analogue of nearly-neutral substitutions in
population genetics \citep{ohta1973slightly}. \emph{Reused} selection sets, which are the common practice, change the
dynamics, and the propositions above no longer describe them. A measured gain's error equals the candidate's score
error minus the incumbent's. The incumbent's positive selection error therefore enters every measured gain with a
negative sign. If later candidates do not inherit an equally large error, this acts as an implicit threshold. We call
this possible mechanism \emph{lock-in}; the fresh-set model does not identify its net size on reused items.
It may make greedy more conservative, possibly at the price of keeping an incumbent chosen partly by luck.
Section~\ref{sec:lockin} measures incumbent inflation and subsequent commit errors, without isolating the mechanism
causally. Appendix~\ref{app:proofs} adds a heuristic, local resolution scale for a given $n$.

\section{Acceptance rules}
\label{sec:method}

All rules choose the best-measured candidate (or, for the Bayes gates, the largest posterior mean) and differ in when
they commit it. \emph{Greedy}: $\hat\Delta_{\hat k}>0$. \emph{McNemar}: a one-sided exact test against the incumbent
at level $\alpha$. \emph{PACE}: the anytime-valid test of \citet{pace2026} as published, which bets half its wealth on
each discordant item and commits when the wealth reaches $1/\alpha$. \emph{Select-then-confirm}: choose on one half of
$D$ and commit iff the winner also beats the incumbent on the other. \emph{Tuned baselines}: McNemar's $\alpha$, or a
threshold on $\hat\Delta_{\hat k}$, chosen to maximize value on pilot runs. \emph{Bayes gates} apply
Equation~\eqref{eq:post} with regularized noise terms from the current generation's item vectors,
$\hat e_\eta^2=\hat v_{cc}/n$ and $\hat e_c^2=(\hat v-\hat v_{cc})/n$, and commit iff the largest posterior mean and
its measured gain are both positive. The \emph{pilot-calibrated} gate takes $(\mu,s^2)$ from a pilot run in which
candidates were also scored on a large evaluation set, separately for the first generation and later ones; the
\emph{online} gate estimates them from the loop's own history. Appendix~\ref{app:alloc} gives the details. The
level of a McNemar or PACE test holds for one fixed candidate on fresh items. Choosing the best of $K$ candidates
before testing and reusing the items across generations both void that guarantee, so we compare rules by the held-out
value of their decisions rather than by their nominal error rates.

\section{Experiments}
\label{sec:exp}

We report an exploratory study, a pre-registered confirmatory study, and studies with a current model, competent
starting points and two prompt optimizers.

\paragraph{Setup.}
Unless stated otherwise, solver and proposer are the same model, Qwen2.5-1.5B- or 7B-Instruct
\citep{qwen2025qwen25}; current-generation models \citep{yang2025qwen3,qwen2026models} appear in
Section~\ref{sec:gepa}. All models are served with vLLM \citep{kwon2023vllm}. In each generation the proposer sees the
current instruction, a task description and four failures, and writes $K=4$ rewrites (Appendix~\ref{app:alloc}). The
tasks are TREC-50 \citep{li2002trec}, Banking77 \citep{casanueva2020banking77} and GSM8K \citep{cobbe2021gsm8k}, and
the seed instructions are one sentence long. Each task has a 600-item \emph{gold} set excluded from selection and feedback
by item ID (the TREC text-overlap audit is in Appendix~\ref{app:alloc}),
and the selection set of $n$ items is reused across generations, as in most systems. We call gold accuracy
\emph{held-out} and the loop's own selection-set score the \emph{proxy}; the proxy minus held-out accuracy is the
inflation of Section~\ref{sec:intro}. A candidate's \emph{held-out effect}, its paired gain on the gold set, estimates
its true effect $\Delta_k$ with a standard error of 1.5 to 2 points. The exploratory runs use $T=20$ generations
and three seeds per configuration, and are replicated with a second inference stack in Appendix~\ref{app:repl}; in its
\emph{oracle-instrumented} runs every candidate is also scored on the gold set.

\subsection{Proposals are mostly harmful, and the winner's curse has the implied size}
\label{sec:measure}

The oracle runs yield 55 to 152 candidates per setting with measured and held-out effects. The first rewrite of a
one-sentence seed often improves on it (on TREC, 75 to 100\% of first-generation candidates do). After that, only 4
to 39\% of rewrites improve on their incumbent, mean held-out effects lie between $-0.2$ and $-13.5$ points, and
occasional catastrophic rewrites make effects left-skewed (excess kurtosis up to 7.6). Stronger proposers
(Qwen2.5-14B, Qwen3.5-4B) change these figures little (Appendix Table~\ref{tab:proposals}). Candidates resemble one
another more than the incumbent, with $v_{cc}\approx0.44$--$0.60\,v$.

\begin{figure}[t]
\centering
\includegraphics[width=\textwidth]{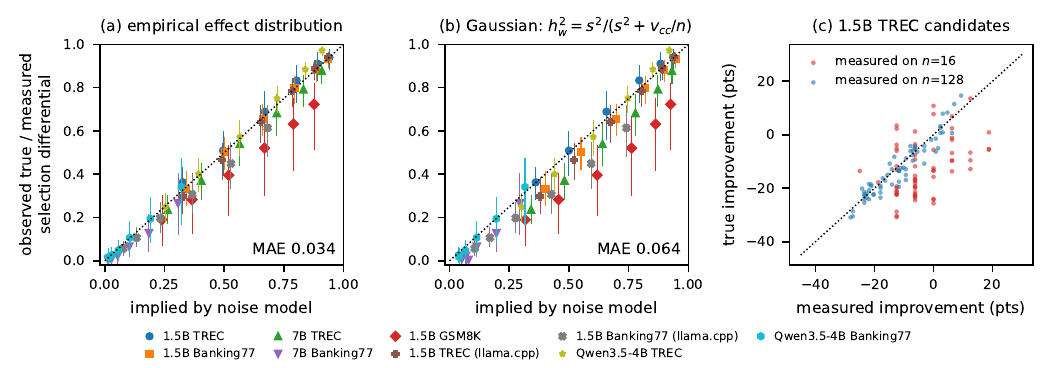}
\caption{\textbf{The winner's curse has the size implied by the noise model.} Observed ratio of the true to the
measured selection differential of the best-of-$K$ rewrite (true: on the held-out part of the split; 95\%
cluster-bootstrap intervals over generations, which omit run-to-run variation) against the ratio implied by (a) the
empirical distribution of held-out effects ($K=4$ approximation) and (b) the Gaussian model, for nine settings and $n\in\{8,\dots,256\}$.
(c) Measured versus held-out effects of the 1.5B TREC candidates on 16 and on 128 items.}
\label{fig:breeder}
\end{figure}

To check the noise model we split the gold set into a pseudo selection set of $n$ items and a truth set, select the
best candidate of each generation on the former, and compare its selection differential on both parts (200 splits;
Figure~\ref{fig:breeder}). With inputs from the same oracle data, the ratio of held-out to measured selection
differential implied by the model matches the observed ratio across nine settings (three models, two Qwen
generations, two inference stacks) and $n=8$ to $256$, with a mean absolute error of 0.034 using the empirical
distribution of held-out effects and 0.064 using the Gaussian $h_w^2$, which overstates reliability where effects are
heavy-tailed. At $n=16$ the winner's held-out selection differential is only 1 to 51\% of its measured
one. A loop can estimate $h_w^2$ itself, with $\hat v_{cc}$ from the candidates' item vectors and $\hat s^2$ from the
within-generation spread of measured effects minus $\hat v_{cc}/n$. Computed causally from generations 1 to $t$ of
one oracle run on a fixed $n$-item subset of its selection set, this estimate predicts the ratio with a mean absolute
error of 0.10 and a bias of $+0.02$, and shrinking the winners' measured advantages by it brings their mean within
0.6 points of the held-out mean (5.0 points unshrunk at $n=16$). A single run's estimate is imprecise (standard
deviation 0.12 to 0.21 at $n=16$; Appendix~\ref{app:online}). Both checks use trajectories that loops with 256 items
generated; Section~\ref{sec:lockin} turns to native small-set loops.

\subsection{Commits on a reused selection set}
\label{sec:lockin}

We compared the model with native loops in a post hoc analysis of the greedy runs of the confirmatory study below
(four settings, eight seeds, $n=16$, 64, 256). For each commit we computed the overstatement that
Assumption~\ref{ass:gauss} implies given acceptance, using the prior $(\mu,s^2)$ of a separate oracle-instrumented
pilot, frozen before those runs, and the noise terms of that generation's candidates on the loop's own selection set
(Appendix Tables~\ref{tab:commits} and~\ref{tab:commits_setting}). The model describes only commits made in
generation 1, when the incumbent is the seed and no earlier decision has used the selection set. For these 72
commits the observed and model averages are 9.1 and 9.0 points at $n=16$, 3.2 and 3.8 at $n=64$, and 0.1 and 1.3 at
$n=256$. The averages hide errors of several points in single
settings (at $n=16$, 1.9 against 5.9 on 1.5B TREC and 12.9 against 7.0 on Banking77, with four to six commits each),
so we read this as agreement on average, not as calibration. For other commits the model value is only a fresh-set
reference. First commits that follow earlier rejections (22 of 94) overstate less than it at $n=16$ (6.3 against
9.6 points), and commits against a selected incumbent overstate less at every $n$ (5.2 against 9.7, 1.9 against 3.7
and 0.5 against 1.1). This pattern is consistent with lock-in: a commit's overstatement is exactly the change in
incumbent inflation, but the net effect depends on how much error later candidates share with the incumbent.

The inflation, relative to the seed's own offset, rises over the first generations and shrinks with $n$. After 11 to
20 generations it is $13.6\pm1.7$ points at $n=16$, $7.2\pm1.1$ at $n=64$ and $1.7\pm0.6$ at $n=256$ (confirmatory
greedy runs, a secondary analysis; Appendix Table~\ref{tab:lockin}). Whether incumbent inflation acts as a
threshold that costs real improvements is less clear. Of the late candidates, 14\% have held-out effects between
0 and 1.8 points, within one gold-set standard error of zero, and almost none clears the incumbent. Re-drawing the
selection set every generation avoids carrying the incumbent's earlier selection error into the next comparison
but, in the one setting where we tried it (three seeds, exploratory design), did not raise the final gain
(Appendix~\ref{app:more}).

\subsection{Pre-registered confirmatory study}
\label{sec:v2}

\begin{figure}[t]
\centering
\includegraphics[width=\textwidth]{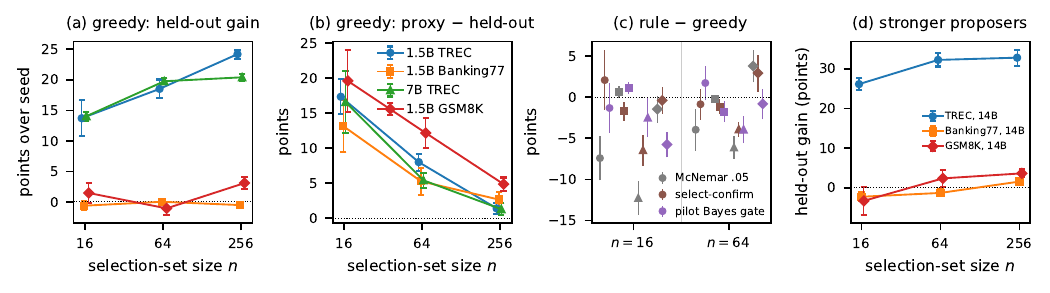}
\caption{\textbf{Pre-registered confirmatory runs} (fresh gold set, eight fresh seeds in (a)--(c) and five in (d), proposer
feedback independent of $n$; mean $\pm$ s.e.). (a) Final held-out gain of greedy loops over the seed instruction and (b) their final
proxy--held-out gap, against $n$. (c) Held-out gain of McNemar ($\alpha=0.05$), select-then-confirm and the
pilot-calibrated Bayes gate minus that of greedy, per setting (paired by seed; settings marked as in (a), legend in (b); colors denote rules). (d) Greedy loops in
which a stronger model proposes for the 1.5B solver.}
\label{fig:v2}
\end{figure}

The exploratory runs had three seeds, and the proposer's failure examples came from the selection set itself, so
that $n$ changed both the evaluation and the information given to the proposer. We therefore pre-registered a
confirmatory study. The design, five hypotheses and an amendment adding select-then-confirm, a phase-indexed pilot
prior and a stronger proposer were fixed in a dated written plan before the first run started. The
analysis script was written and archived ten minutes into the runs, when 16 of 336 had finished. Deviations from the plan
are listed in Appendix~\ref{app:alloc}. The gold set is a fresh 600-item sample disjoint from the exploratory one, the seeds are new, and the
proposer's failure examples come from a separate pool of 256 items, so that $n$ changes only the evaluation. Four
settings (1.5B TREC, 1.5B Banking77, 7B TREC, 1.5B GSM8K) each have nine arms with eight seeds, namely greedy
acceptance at $n=16,64,256$ and McNemar at $\alpha=0.05$, the pilot-calibrated Bayes gate and select-then-confirm at
$n=16$ and $64$. The pilot prior was frozen before launch from the exploratory oracle runs of the same model and task. Tests are
paired by seed, two-sided, and Holm-corrected within each family. Table~\ref{tab:v2} and Figure~\ref{fig:v2} report
the results, and Appendix~\ref{app:more} lists all arms.

\begin{table}[t]
\centering
\caption{\textbf{Pre-registered tests} (points; mean $\pm$ s.e.\ of seed-paired differences; $p$ Holm-adjusted
within each family, unadjusted for the pooled rows).}
\label{tab:v2}
\small
\setlength{\tabcolsep}{4pt}
\begin{tabular}{lccc}
\toprule
Setting & difference (pts) & wins / pairs & $p$ \\
\midrule
\multicolumn{4}{l}{\emph{H1: gain($n{=}256$) $-$ gain($n{=}16$), greedy}} \\
1.5B TREC & +10.5$\pm$2.6 & 7 / 8 & 0.010 \\
7B TREC & +6.4$\pm$0.7 & 8 / 8 & $<$0.001 \\
1.5B GSM8K & +1.6$\pm$2.2 & 4 / 8 & 0.483 \\
\midrule
\multicolumn{4}{l}{\emph{H2: gap($n{=}16$) $-$ gap($n{=}256$), greedy}} \\
1.5B TREC & +15.9$\pm$2.7 & 8 / 8 & 0.002 \\
1.5B Banking77 & +10.5$\pm$4.6 & 7 / 8 & 0.058 \\
7B TREC & +15.3$\pm$4.7 & 7 / 8 & 0.037 \\
1.5B GSM8K & +14.7$\pm$4.4 & 8 / 8 & 0.037 \\
\midrule
\multicolumn{4}{l}{\emph{H3: greedy gain at $n{=}16$ ($t$-test; sign test)}} \\
1.5B Banking77 & -0.6$\pm$0.7 & 4 lose / 8 & 0.425; 1.000 \\
\midrule
\multicolumn{4}{l}{\emph{H4: pilot gate $-$ greedy at $n{=}16$}} \\
pooled, 32 pairs & -2.1$\pm$1.1 & 10 / 32 & 0.062 \\
1.5B TREC & -1.3$\pm$3.0 & 3 / 8 & 0.677 \\
1.5B Banking77 & +1.1$\pm$0.7 & 4 / 8 & 0.552 \\
7B TREC & -2.5$\pm$2.4 & 2 / 8 & 0.650 \\
1.5B GSM8K & -5.7$\pm$1.6 & 1 / 8 & 0.032 \\
\midrule
\multicolumn{4}{l}{\emph{H5: select-confirm $-$ greedy at $n{=}16$}} \\
pooled, 32 pairs & -1.6$\pm$1.2 & 9 / 32 & 0.194 \\
1.5B TREC & +2.1$\pm$3.6 & 5 / 8 & 1.000 \\
1.5B Banking77 & -1.7$\pm$1.2 & 1 / 8 & 0.566 \\
7B TREC & -6.4$\pm$1.9 & 0 / 8 & 0.045 \\
1.5B GSM8K & -0.4$\pm$1.6 & 3 / 8 & 1.000 \\
\bottomrule
\end{tabular}

\end{table}

The clearest result concerns the loops' reports about themselves (H2). With 16 items, the final proxy exceeded
held-out accuracy by 13 to 20 points, and with 256 items by 1 to 5 points. The paired reduction was about 15 points (14.7 to
15.9) in three settings (Holm-adjusted $p\le0.04$) and 10.5 points on Banking77, where it fell short of significance
($p=0.058$). Held-out gains rose with $n$ on TREC (H1), from $13.7\pm2.9$ to $24.2\pm0.7$
points for the 1.5B model (seven of eight seeds) and from $14.0\pm0.7$ to $20.4\pm0.6$ for the 7B model (eight of
eight). They did not rise on GSM8K, where no arm gained more than 3.1 points. The Goodhart loss that the
exploratory runs showed on Banking77 did not replicate (H3). Greedy loops at $n=16$ ended $0.6\pm0.7$ points below the
seed, with four of eight seeds losing, while their selection-set scores rose by 12.5 points. The exploratory loss was
itself not significant ($p=0.20$, three seeds).

The tested acceptance rules did not improve whole runs (H4, H5). At $n=16$ the pilot-calibrated gate ended
$2.1\pm1.1$ points below greedy, pooled over 32 seed pairs ($p=0.06$; 95\% $t$-interval $[-4.3,0.1]$), and it was
significantly worse on GSM8K, whose pilot prior rests on a single oracle run. Select-then-confirm ended $1.6\pm1.2$ points below greedy. Fixed-level McNemar tests gave up most of the
available gain on TREC ($-7.4$ and $-12.3$ points for the 1.5B and 7B models). When Qwen2.5-14B proposes for the 1.5B
solver, TREC gains rise to 26 to 33 points, and the self-reported gap still shrinks significantly with $n$ on all three
tasks (Figure~\ref{fig:v2}d).

\begin{figure}[t]
\centering
\includegraphics[width=\textwidth]{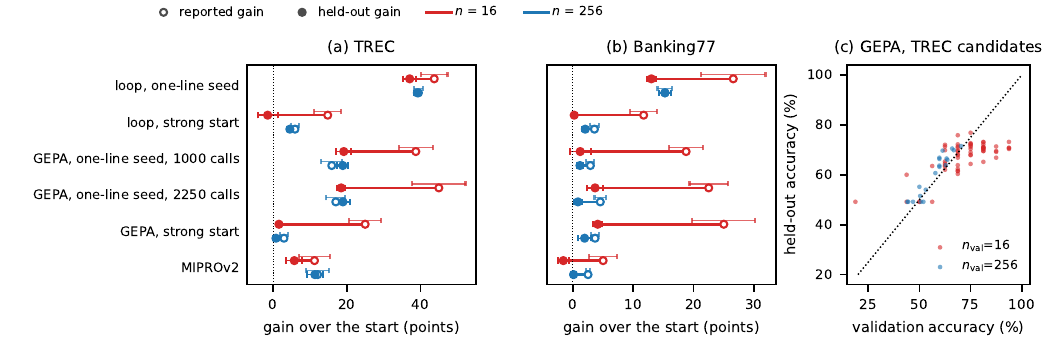}
\caption{\textbf{What loops and optimizers report, and what they gain.} (a, b) Gain of the final incumbent or returned
program over the starting instruction or program, as measured on the selection (validation) set (open markers) and on
600 held-out items (filled), with 16 (red) or 256 (blue) selection items; the segment is the overstatement. Loops:
Qwen3.5-4B improving its own instruction from the one-sentence seed or from a competent start (20 generations). GEPA:
from the seed program at 1000 and 2250 metric calls, and from a strong program at 2750 (TREC) and 2000 (Banking77)
calls. MIPROv2: 24 trials. Means over four to eight seeds, with $\pm1$ s.e.\ (held-out on
the segment, reported above it). (c) Validation and held-out
accuracy of the held-out-scored TREC candidates of GEPA at 1000 calls.}
\label{fig:systems}
\end{figure}

\subsection{Current models, strong starts and two prompt optimizers}
\label{sec:gepa}

We first repeated the loop with Qwen3.5-4B as its own proposer and solver (four seeds; Appendix~\ref{app:alloc}).
From the one-sentence seed it gained 37 to 39 points on TREC and 13 to 15 on Banking77 at every $n$, and the gap again
shrank with $n$ (Banking77: 15.7 to 3.4 points, $p=0.007$). These loops amount to a single rewrite: the first
generation delivers 90 to 112\% of the final gain, and only 7 and 20\% of later rewrites improve on their incumbent.
We therefore restarted the loop from a competent instruction, the final instruction of one of its own earlier runs,
chosen before any run. From there only 5\% of TREC rewrites improve (mean effect $-14$ points), and 45\% of Banking77
rewrites do, but only 5\% by more than 2 points (Appendix Table~\ref{tab:strongstart}). With the four planned seeds,
loops with 16 items reported gains of 12.5 and 14.1 points on TREC and Banking77 while held-out accuracy changed by
$-2.2$ and 0.0; with 256 items they reported 5.4 and 3.3 and gained 4.3 and 1.8. The planned proxy--held-out gap fell
by 9.4 and 2.8 points (two-sided $p=0.12$ and 0.69). Four seeds added after seeing these results gave gap reductions
of 7.2 and 12.9 points ($p=0.08$ and 0.03) and held-out advantages of the larger set of 5.6 and 1.9 ($p=0.15$ and
0.16). Over eight seeds the overstatement, a measure we added because the start's offset varies across 16-item sets,
fell from 16.3 to 1.4 and from 11.5 to 1.5 points. We decided on the extension after seeing the first stage and
chose the one-sided direction and equal stage weights after both stages had run, so the combined inverse-normal
$p$-values (0.010 and 0.033 for the gap, 0.04
for the held-out advantage, 0.004 and $<0.001$ for the overstatement) are exploratory.

We also instrumented two prompt optimizers in DSPy \citep{khattab2024dspy} with the same held-out set. GEPA
\citep{agrawal2025gepa} evolves a program's instruction by reflecting on failures in training minibatches and returns
the candidate with the best score on a validation set that it reuses throughout the run. MIPROv2
\citep{opsahlong2024mipro} proposes a fixed set of candidate instructions and searches among them by Bayesian
optimization on the validation set. The GEPA paper reports accuracy on separate test sets, so what we measure is the
optimism of the validation scores on which the optimizers select, not of GEPA's published gains. The task model is Qwen3.5-4B and the proposer Qwen3.8-27B, both without thinking,
with a 256-item training pool, $n_{\mathrm{val}}$ validation items and five seeds (Appendix~\ref{app:alloc}). Because
a candidate's validation scores are fixed when GEPA finds it, the program returned after any number of metric calls
can be reconstructed exactly from a run's saved state or log; we checked every reconstruction against the logs.
Appendix~\ref{app:alloc} lists the changes to the recorded plan, including the reported-gain measure, added after
eight runs with $n_{\mathrm{val}}=256$ but none with fewer items had been seen.

Figure~\ref{fig:systems} collects the results. At 1000 metric calls GEPA's returned program gained about 19 points of
held-out accuracy on TREC, 1 to 2 on Banking77 and nothing on GSM8K, where the task model already scores 95\%, whatever
$n_{\mathrm{val}}$ was. The validation score on which GEPA selects gave a different picture. With 16 items it put the
improvement at 38.8 points on TREC for a real 19.2, at 18.8 on Banking77 for a real $1.2\pm1.7$, and at 6.2 on GSM8K
for a program 0.9 points worse. With 256 items the two differed by at most 3 points (Appendix Table~\ref{tab:gepa}). At that budget, however, the 256-item arm had found three candidates and made a tenth of the reflections
of the 16-item arm. Run to 2250 calls, where its pool holds 9 programs and that of the 16-item arm 91, GEPA's held-out
gain still did not grow with $n_{\mathrm{val}}$ (TREC: 18.5, 20.4 and 19.0 points with 16, 64 and 256 items;
Banking77: 3.7, 1.3 and 0.8). The overstatement relative to the starting program remained large with 16 items, 26.5
points against $-1.9$ on TREC ($p=0.007$) and 18.8 against 3.7 on Banking77 ($p=0.002$). Started from its own best
program, GEPA with 16 validation items reported improvements of 25.0 points on both tasks, and the real ones were 1.6
and 4.1 (256 items: 3.0 and 3.7 reported, 0.8 and 1.9 real). MIPROv2 largely holds the proposed instruction pool
fixed across its two arms. With 256 items its returned program gained 11.4 held-out points on TREC
against 5.8 with 16 (paired difference $5.6\pm0.7$, $p=0.001$, all five seeds). On Banking77 its 16-item arm returned
programs that were 1.6 points worse than the seed, although their validation scores implied a 5-point gain (overstatement 6.6 against 2.4
points with 256 items, $p=0.27$; Appendix~\ref{app:mipro}). Proposition~\ref{prop:breeder}, with $K$ the pool size
and noise terms from the validation item vectors that each optimizer records, tracks how the overstatement varies
with $n_{\mathrm{val}}$ and budget but is not calibrated. Where GEPA's overstatement is large, the observed value is
1.1 to 1.9 times the prediction (GEPA's pool is itself selected on the validation set), and the model overpredicts
MIPROv2's 16-item TREC cell (9.0 against 5.5 points; Appendix~\ref{app:mipro}).

\section{Discussion and limitations}
\label{sec:disc}

\paragraph{What to log and report.}
Our most robust finding is that a loop's own estimate of its progress is too optimistic, in the Qwen loops and in the
validation scores of GEPA and MIPROv2 alike. The loop's item-level logs support an estimate of how much a
candidate's advantage over its generation's mean should shrink (Section~\ref{sec:measure}). Predicting how much an
accepted gain is overstated also uses a prior over effects, which our native-loop analysis took from a separate pilot
(Section~\ref{sec:lockin}), and neither identifies the inflation that accumulates on a reused selection set.
Shrinking each commit by the loop's own $\hat h_w^2$ largely removes the average
overstatement of the reported gain across our 151 confirmatory and current-model runs with $n=16$, but leaves $-8.5$
to $+5.8$ points within a setting, because that error is shared by all candidates. A small audit set does estimate
it. Scoring the seed and the incumbent after 19 generations on 64 items never used for selection costs 128
evaluations, a tenth of the nominal budget of 1{,}280 candidate evaluations of a loop with $n=16$. It removes the
bias of the reported gain and reduces its root-mean-square error from 15.5 to 5.7 points. That error is still as
large as a typical gain, and at the disagreement rates of our tasks, resolving gains of a few points would take a few
hundred items (Appendix~\ref{app:online}). We recommend that
self-improvement studies report a held-out score and its standard error alongside the score used for selection.

\paragraph{What acceptance rules can and cannot do.}
In our experiments, calibrated thresholds improved isolated decisions on fresh evaluations, mainly by committing
less often (Appendix~\ref{app:more}), but the tested rules did not improve complete runs on reused selection sets.
Incumbent inflation may make greedy more conservative when later candidates do not share equally large errors.
A larger selection set raised held-out gains on TREC in the Qwen2.5 runs and in
MIPROv2 at sixteen times the evaluation cost, but not detectably in GEPA, where the small set allowed ten times more
search. We therefore do not recommend the tested rules as a general remedy for adaptive reuse.

\paragraph{Limitations.}
Our artifacts are instructions, our tasks use deterministic scoring rules (though model outputs are not bitwise
reproducible; Appendix~\ref{app:alloc}), our largest proposer has 27B
parameters, and we have not examined agentic or code tasks. Stochastic rollouts introduce additional measurement
noise; to the extent that it is candidate-specific, Proposition~\ref{prop:breeder} predicts a larger winner's curse,
holding the effect distribution fixed. The confirmatory study uses Qwen2.5 models, as
pre-registered. Both checks in Figure~\ref{fig:breeder} use oracle quantities, the native-loop check of
Section~\ref{sec:lockin} covers four Qwen2.5 settings and is post hoc, the Bayes gates need a pilot with a large
evaluation set, and PACE ran with its published fixed bet and $n$ items. We do not model the dynamics of lock-in or
make claims about recursive self-improvement.

\section*{Code and data availability}
The code, saved run records, and reproduction instructions are provided in the ancillary files accompanying
this preprint. Appendix~\ref{app:alloc} describes the experimental designs and provenance.

\bibliographystyle{apalike}
\bibliography{refs}

\clearpage
\appendix
\section{Proofs}
\label{app:proofs}

Throughout, $\phi,\Phi$ are the standard normal density and CDF, and $Z_{(K)}=\max_{k\le K}Z_k$ for i.i.d.\
standard normals.

\paragraph{Lemma A.1 (posterior under the random-effects model).}
Let $\hat\Delta=\Delta+c\mathbf{1}+\eta$ with $\Delta_k\stackrel{iid}{\sim}N(\mu,s^2)$, $c\sim N(0,e_c^2)$ and
$\eta_k\stackrel{iid}{\sim}N(0,e_\eta^2)$ independent. Then $\mathrm{Cov}(\Delta_k,\hat\Delta)=s^2e_k$ and
$\Sigma=\mathrm{Var}(\hat\Delta)=aI+e_c^2\mathbf{1}\mathbf{1}^\top$ with $a=\sigma_w^2=s^2+e_\eta^2$. By
Sherman--Morrison, $\Sigma^{-1}=a^{-1}\big(I-\tfrac{e_c^2}{a+Ke_c^2}\mathbf{1}\mathbf{1}^\top\big)$, so
\[
\mathbb{E}[\Delta_k\mid\hat\Delta]=\mu+s^2e_k^\top\Sigma^{-1}(\hat\Delta-\mu\mathbf{1})
=\mu+h_w^2(\hat\Delta_k-\bar m)+\frac{h_w^2a}{a+Ke_c^2}(\bar m-\mu),
\]
which is Equation~\eqref{eq:post}. Averaging over $k$ gives
$\mathbb{E}[\bar\Delta\mid\hat\Delta]=\mu+\frac{h_w^2a}{a+Ke_c^2}(\bar m-\mu)$, hence
$\mathbb{E}[\Delta_k-\bar\Delta\mid\hat\Delta]=h_w^2(\hat\Delta_k-\bar m)$. $\square$

\paragraph{Proof of Proposition~\ref{prop:breeder}.}
$\hat k$ is a function of $\hat\Delta$, so Lemma A.1 and the tower property give
$\mathbb{E}[\Delta_{\hat k}-\bar\Delta\mid\hat\Delta]=h_w^2(\hat\Delta_{\hat k}-\bar m)$. The within-generation
deviations $X_k=\hat\Delta_k-c=\Delta_k+\eta_k$ are i.i.d.\ $N(\mu,\sigma_w^2)$ and $\hat\Delta_k-\bar m=X_k-\bar X$, so
$\mathbb{E}[\hat\Delta_{\hat k}-\bar m]=\mathbb{E}\max_kX_k-\mathbb{E}\bar X=\sigma_w\kappa_K$. For the overstatement,
$\hat\Delta_{\hat k}-\Delta_{\hat k}=(\hat\Delta_{\hat k}-\bar m)-(\Delta_{\hat k}-\bar\Delta)+(\bar m-\bar\Delta)$ and
$\bar m-\bar\Delta=c+\bar\eta$; taking expectations gives $(1-h_w^2)\sigma_w\kappa_K+\mathbb{E}[c]$. With $e_c=0$,
$h_w^2=h^2$ and the first statement is the breeder's equation. $\square$

\paragraph{Proof of Proposition~\ref{prop:bayes}.}
For any rule $A(\hat\Delta)\in\{1,\dots,K,\varnothing\}$ with $\Delta_\varnothing=0$,
$\mathbb{E}\Delta_A=\mathbb{E}\big[\sum_k1\{A=k\}\mathbb{E}[\Delta_k\mid\hat\Delta]\big]$, maximized pointwise by
choosing the largest posterior mean if positive and $\varnothing$ otherwise. The posterior mean is increasing in
$\hat\Delta_k$ (the other terms are common to all $k$), so the largest belongs to $\hat k$. With $e_c=0$,
$\mathbb{E}[\Delta_k\mid\hat\Delta_k]=\mu+h^2(\hat\Delta_k-\mu)>0$ iff $\hat\Delta_k>\mu-\mu/h^2=-\mu e^2/s^2$
when $s>0$. When $s=0$, every true effect equals the known $\mu$, so commit iff $\mu>0$, independently of the observations. $\square$

\paragraph{Proof of Proposition~\ref{prop:goodhart}.}
With $s=0$, $\hat\Delta_k=\mu+c+e_\eta Z_k$ and greedy commits iff $c+e_\eta\max_kZ_k>|\mu|$. This probability
increases with $K$ (the maximum increases) and decreases with $n$ ($e_c^2$ and $e_\eta^2$ scale as $1/n$). Each commit
changes $f$ by exactly $\mu$, and commits require $\hat\Delta_{\hat k}>0$. With fresh selection sets and proposal and
noise distributions that stay fixed, the events are independent and identically distributed across generations,
giving expected drift $\mu p$ per generation. The statement is local to such a stationary regime. $\square$

\paragraph{A heuristic resolution scale (Section~\ref{sec:model}).}
Let the proposal distribution depend on the incumbent's performance $f$, with $b(f)=-\mu/s>0$ (harmful proposals on
average). With independent noise and the rule of Proposition~\ref{prop:bayes}, the expected one-step gain at a state
with parameters $(s,b,e)$ is $G^\star=s\,\psi_{K,b}(s/e)$ with $\psi_{K,b}(r)=\mathbb{E}[(\rho Z_{(K)}-b)^+]$ and
$\rho=r/\sqrt{1+r^2}$. This gain is negligible at states where the standardized threshold
$\lambda=b/\rho=-\mu\sqrt{s^2+e^2}/s^2$ exceeds $\sqrt{2\ln(KT)}$: even $T$ such steps add less than a small multiple
of $s$. We use the level at which $\lambda$ reaches that value only as a qualitative, local resolution scale. When
$\lambda$ increases with performance near the crossing, this level rises with $n$ in the noise-dominated regime
($v/n\gtrsim s^2$). It depends on $K$ and $T$ only through $\sqrt{2\ln(KT)}$, and as
$n\to\infty$ is set by the proposal distribution alone, in analogy to the drift barrier \citep{sung2012drift} and to
progress limits of noisy evolution strategies \citep{arnold2006general}. It is not a bound on the level a loop attains
over $T$ generations. Such a bound would need uniform control over all reachable states, since $s$, $b$ and $\rho$
change with the incumbent and accuracy can fall as well as rise. The derivation of the one-step statement follows.
With $e_c=0$, $\mu=-bs$ and the rule of Proposition~\ref{prop:bayes}, the gain of one generation is
$\mathbb{E}[(\mu+h^2(\hat\Delta_{\hat k}-\mu))^+]$. With $\hat\Delta_{\hat k}=\mu+\sigma Z_{(K)}$ and
$h^2\sigma=s\rho$, $\rho=s/\sigma=r/\sqrt{1+r^2}$, this is $s\,\mathbb{E}[(\rho Z_{(K)}-b)^+]=s\psi_{K,b}(r)$. Since
$(\rho Z_{(K)}-b)^+\le\sum_k(\rho Z_k-b)^+$, with $t=b/\rho$,
$\psi_{K,b}\le K\rho[\phi(t)-t(1-\Phi(t))]\le K\rho\phi(t)/t^2=K\rho^3\phi(b/\rho)/b^2$ by the Mills-ratio bound
$1-\Phi(t)\ge\phi(t)(1/t-1/t^3)$, valid for $t>0$. At a state with $b/\rho\ge\lambda_0$ the expected one-step gain is
therefore at most $Ks\rho^3e^{-\lambda_0^2/2}/(\sqrt{2\pi}b^2)$, and $T$ gains of this size sum to less than
$s\rho^3/(\sqrt{2\pi}b^2)$ once $\lambda_0\ge\sqrt{2\ln(KT)}$. Special cases: if $b$ is constant and
$s=a(f^\star_\infty-f)^\beta$ with $a,\beta>0$, the level at which $\lambda$ reaches $\sqrt{2\ln(KT)}$ satisfies
$f^\star_\infty-f^\star_n\approx\big(b^2v/(2a^2n\ln(KT))\big)^{1/(2\beta)}$ in the noise-dominated regime; if $s$ is
constant, $b(f^\star_\infty)=\sqrt{2\ln(KT)}$, and $b$ is differentiable near $f^\star_\infty$ with
$0<b'(f^\star_\infty)<\infty$, local inversion gives
$f^\star_\infty-f^\star_n\propto(1-\rho(n))\approx v/(2ns^2)$.

\paragraph{Remark (what the model leaves out).}
(i) Candidate noise is heteroscedastic ($v_k$ depends on both the effect and the candidate--incumbent disagreement
rate) and true effects can be heavy-tailed; the
distribution-aware prediction of Section~\ref{sec:exp} replaces the Gaussian prior by the empirical one. (ii) On a
reused selection set the shared error may have non-zero mean, and proposals informed by failures on $D$ can adapt
to $D$ itself. Incumbent inflation alone does not identify the net gain error; Section~\ref{sec:lockin} reports
patterns consistent with lock-in. (iii) Stochastic per-item rewards (agentic tasks) add a within-item
variance term to $v$.

\section{Experimental details}
\label{app:alloc}

\subsection{Loop, prompts and rules}
\paragraph{Solver prompt.} The artifact is the system message; the user message is fixed:
\begin{quote}\small\ttfamily\raggedright
Possible labels: ABBR:abb, ABBR:exp, \dots\ (all labels)\\[2pt]
Question: \{question\}\\ Label:
\end{quote}
(Banking77: ``Possible intents'' / ``Customer message'' / ``Intent''; GSM8K: ``Question: \{question\}''.) Decoding is
greedy (48 new tokens for classification, 320 for GSM8K). A classification prediction is the first line if it matches
a label, otherwise the last label string in the output; a GSM8K prediction is the last number following
``\#\#\#\#'', then the last number following ``answer is'' if no such marker is present, otherwise the last number.

\paragraph{Proposer prompt.} The proposer (the solver itself unless stated otherwise) receives
\begin{quote}\small\ttfamily\raggedright
A small language model is given the following INSTRUCTION as its system prompt and then solves one task
input. TASK: \{task description\}. CURRENT INSTRUCTION: <<<\{artifact\}>>>. Here are inputs where the model
made mistakes with the current instruction: \{4 failures: input, model output, correct answer\}. Write an
improved INSTRUCTION that will make the model more accurate on this task in general (not only on these
examples). You may add strategies, rules, definitions, or output-format requirements. Keep it under 120
words. Return only the new instruction between <<< and >>>.
\end{quote}
with system message ``You are an expert prompt engineer who improves instructions for a smaller language model'',
temperature 0.9, top-$p$ 0.95, at most 400 new tokens. A proposal is valid if it contains a complete block opened by
\texttt{<<<} and closed by \texttt{>>>} (or a second \texttt{<<<}) of at most 180 words; invalid proposals and
duplicates are replaced by fresh samples (3 rounds in the exploratory runs, 5 in the confirmatory runs). Even so,
some generations have fewer than the intended $K$ valid candidates: 26.0\% (1.5B, vLLM), 25.8\% (GSM8K), 0.0\% (7B)
and 39.3\% (1.5B, llama.cpp) in the exploratory runs, 6.6\% in the confirmatory runs, and almost none with the 14B and Qwen3.5-4B
proposers (per arm in Table~\ref{tab:v2arms}). In the exploratory runs failures are sampled from the selection set; in the
confirmatory runs from a separate feedback pool of 256 items (independent of $n$).

\paragraph{Seed instructions and task descriptions.} The seed instructions are ``Classify the question.'' (TREC),
``Classify the customer message.'' (Banking77) and ``Answer the question.'' (GSM8K). The task descriptions shown to the
proposer are: TREC, ``Classify a question by the type of answer it seeks into one of 50 fine-grained TREC labels of
the form COARSE:fine (the label list is given in the input). The first line of the response is matched against the
label set; otherwise the last label mentioned is used.''; Banking77, ``Classify an online-banking customer message
into one of 77 intent labels (the label list is given in the input). The first line of the response is matched
against the label set; otherwise the last label mentioned anywhere in the response is used.''; GSM8K, ``Grade-school
math word problems. The final numeric answer is extracted from the response (a number after `\#\#\#\#' if present,
otherwise the last number in the response) and compared to the reference.'' The strong-start loops of
Section~\ref{sec:gepa} start from the following instructions, verbatim (line breaks as in the original).
\begin{quote}\small\ttfamily\raggedright
TREC: Classify question intent using TREC COARSE:FINE labels. Strictly match the fine-grained label to the question
type:\\ 1. Entities: COARSE:ENTY (words, names, objects). If asking for a specific *instance* (e.g., "which novel?",
"what nickname?"), use ENTY:cremat. If asking for an *event occurrence* (e.g., "when did X happen?"), use
ENTY:event.\\ 2. People: HUM:ind for specific individuals/nicknames; HUM:gr for groups/organizations.\\ 3.
Attributes: DESC:def for abstract meanings; DESC:desc for general facts/origin; NUM:date for specific timepoints.\\
4. Rules: Prioritize the question's target entity over descriptive content. Output ONLY the label.\\[3pt]
Banking77: You are an intent classifier. Map the input to exactly one label from the provided list. Prioritize the
MOST SPECIFIC label describing the exact transaction event or specific charge (e.g., 'topping\_up\_by\_card' for card
top-up failures, 'wrong\_exchange\_rate\_for\_cash\_withdrawal' for exchange errors). Avoid broad categories like
'fee charged' if a specific event label exists. Match the core user issue precisely, ignoring filler. If no exact
match, choose the most semantically similar. Output ONLY the label name on the first line. Do not use explanations,
quotes, or markdown.
\end{quote}
The strong-start GEPA programs (755 and 2151 words) are in the supplementary package.

\paragraph{Pools and splits.} A task's pool concatenates the public train and test splits in their published order;
item $i$ of split $s$ has identifier \texttt{task-s-i} (with prefix \texttt{b77} for Banking77; TREC:
\texttt{CogComp/trec}, fine labels; Banking77:
\texttt{PolyAI/banking77}; GSM8K: \texttt{openai/gsm8k}, configuration \texttt{main}; the first two from the
\texttt{refs/convert/parquet} revision). With Python's \texttt{random} module, the exploratory gold set comprises
the first 600 items after shuffling the pool with \texttt{Random(1234)}. Keeping the remaining items in that
shuffled order, we reshuffle them with \texttt{Random(2027)} and take the first 600 as the confirmatory gold set.
The selection pool is the pool without the exploratory gold set in
the exploratory runs and without both gold sets otherwise. In a loop with seed $r$,
\texttt{Random(r)} first samples the 256-item feedback pool (confirmatory design) and then a fixed ordering of the
remaining items, whose first $n$ items are the selection set, so selection sets are nested across $n$. GEPA and
MIPROv2 shuffle the confirmatory selection pool with \texttt{Random(1000+r)} and take the first 256 items as the
training pool and the next $n_{\mathrm{val}}$ as the validation set. The supplementary package lists the item
pool order and recorded split identifiers; optimizer splits whose identifiers were not saved can be reconstructed
from the manifests, task and run seed using the procedure above. Model weights are the Hugging Face releases named in the text
(\texttt{Qwen/Qwen2.5-\{1.5B,7B,14B\}-Instruct}, \texttt{Qwen/Qwen3.5-4B}, \texttt{Qwen/Qwen3.8-27B-FP8}); we did
not pin their revisions.

\paragraph{Duplicate-question audit.} Disjoint item identifiers do not ensure distinct question text. A post hoc
audit found 12 exact TREC question texts shared by the confirmatory gold set and its remaining pool; 58 of the 144
TREC confirmatory runs include a matching question in their selection or feedback set, affecting at most two of
the 600 gold items per run. Removing all 12 matching gold items leaves 588 items and preserves the primary
conclusions: the TREC larger-set advantages are 10.44 and 6.21 points (Holm-adjusted $p=0.010$ and $<0.001$),
and the pooled pilot-versus-greedy comparison is $-2.10\pm1.08$ points ($p=0.061$). The supplementary package
contains the duplicate identifiers and a reproducible sensitivity analysis. No original run record or primary
estimate was replaced. Banking77 has one gold--remaining-pool text duplicate after trimming surrounding whitespace, which appears in none of its
confirmatory selection or feedback sets; GSM8K has none. Thus held-out refers to item-ID separation, with the
documented TREC exception to content separation.
One final item vector missing from the run records was reconstructed from retained cached model outputs; its
aggregate score matches the recorded score, which does not establish itemwise identity under nondeterministic
inference. The sensitivity conclusion also holds for all 13 possible numbers of correct answers among the 12
excluded items for that run, without relying on the reconstruction.

\paragraph{Rule details.} McNemar: one-sided exact test of the best candidate, not corrected for the choice among $K$.
PACE (decision-level analysis): the published test, a fixed bet $\lambda=1/2$ on each discordant item (ties
skipped), commit as soon as the wealth reaches $1/\alpha$ ($\alpha=0.05$) within $n$ items, reject if the items run
out first; it needs at least eight more wins than losses, which a 16-item set rarely provides. Bayes gates use
$\hat v=\overline{\max\{\hat q_k-\hat\Delta_k^2,1/n\}}$ and
$\hat v_{cc}=\min\{\hat v,\tfrac12\overline{\max\{\hat q_{jk}-(\hat\Delta_j-\hat\Delta_k)^2,1/n\}}\}$,
averaging over candidates and candidate pairs, respectively. With one candidate the implementation uses
$\hat v_{cc}=0.6\hat v$. These numerical floors and the cap precede the noise decomposition in
Section~\ref{sec:method}; the native-commit analysis uses unfloored moments instead. The prior is phase-indexed because rewriting a one-line seed is often
beneficial while later rewrites mostly are not. The requirement that the measured gain also be positive was added
after the online gate committed measured regressions on our development seed when its estimate of $s^2$ collapsed
(Appendix~\ref{app:more}). E-process in the
exploratory whole runs: a predictable GRAPA-style bet $\lambda_j=\min\{1/2,\max(0,\bar d_j/(\hat\sigma^2_j+\bar d_j^2))\}$
computed from the preceding items, commit iff the running maximum of the wealth reaches $1/\alpha$. Online EB gate: window of 8 generations; McNemar fallback in the first generation; for $K=1$, $s^2$ from the
across-generation variance minus the total noise. Pilot-calibrated gate: phase-indexed prior (generation 1 vs.\ later)
from the held-out effects of oracle runs of the same model and task, measured on the exploratory gold set (disjoint from the
confirmatory gold set). In points, $(\mu,s)$ for generation 1 and for later generations is $(5.6,8.6)$ and
$(-9.4,7.1)$ for 1.5B TREC, $(-6.8,6.1)$ and $(-4.6,5.1)$ for 1.5B Banking77, $(4.9,4.8)$ and $(-7.2,7.2)$ for 7B
TREC, and $(-7.7,5.1)$ and $(-2.8,7.6)$ for 1.5B GSM8K. One of the three 1.5B TREC source runs (llama.cpp, 20
generations) was still running when the priors were frozen, so its prior uses that run's first 17 generations;
with all 20, the later-generation values would be $(-8.9,7.0)$. \emph{Harmful commit}: a commit whose held-out
effect on the gold set is negative; the gold set resolves paired differences to about 1.5--2 points, so we also report
commits worse than $-1.8$ points.

The frozen pilot estimator subtracted total candidate--incumbent sampling variance from the within-generation
spread. Under Assumption~\ref{ass:gauss}, the shared incumbent error cancels when centering candidates, so this
historical approximation subtracts too much noise. We retain those frozen values because they governed the
completed experiments and the pilot-based commit references. The post hoc noise-model analysis in Figure~\ref{fig:breeder}
and the proposal spreads in Table~\ref{tab:proposals} instead subtract candidate--candidate noise, with weights
matching the pooled within-generation variance. The empirical-distribution predictor resamples four effect
deviations per simulated generation, a fixed-$K$ approximation when an observed generation has fewer valid
candidates. Figure~\ref{fig:breeder}'s observation intervals bootstrap generations from the same 200 splits used
for the plotted observations.

\subsection{Designs}
\paragraph{Exploratory runs (v1).} Gold set: 600 items per task (fixed random subset, seed 1234). Selection sets
reused across generations and nested across $n$ for a given seed. Main stack: vLLM 0.30.0 (bf16) on one rented
A100-SXM4-80GB, $T=20$ generations, 3 seeds. Oracle-instrumented runs (every candidate scored on the gold set) used
greedy acceptance with $n=256$, $K=4$: one run each for 1.5B-TREC, 1.5B-Banking77 and GSM8K ($T=20$, $T=20$, $T=15$) and
two each for 7B-TREC and 7B-Banking77 ($T=15$). Replication stack: llama.cpp (build b11211, 8-bit GGUF) on a shared
RTX~3080~Ti, $T=25$ (two oracle runs per task, $T=20$).

\paragraph{Confirmatory runs (v2).} Pre-registered in a dated written plan before launch: a fresh
gold set of 600 items per task disjoint from the v1 gold set; seeds 11--18; feedback pool of 256 items disjoint from
selection sets and gold; arms greedy ($n=16,64,256$), McNemar, pilot-calibrated Bayes gate and select-then-confirm
($n=16,64$); settings 1.5B-TREC, 1.5B-Banking77, 7B-TREC, 1.5B-GSM8K; stronger proposer: Qwen2.5-14B proposing for
the 1.5B solver (greedy, $n=16,64,256$, seeds 11--15, one oracle run per task). vLLM on one A100 as above.

\paragraph{Deviations from the pre-registration.}
(i) The pre-registration file, including its amendment, and the pilot priors were last modified at 15:54; the first run
started at 16:07. The amendment's own header gives a wrong time (``16:15''). (ii) The analysis script, which implements
the pre-registered tests, was written ten minutes after launch (file created 16:17:46, when 16 of 336 runs had
finished), archived at 16:17:50, and run on those 16 runs at 16:17:52. That run printed the per-arm summary and failed
at the output step because of a serialization error; fixing that error is the only change made to the script. (iii) The pooled tests of H4 and H5 were implemented as one-sample $t$-tests on the 32 seed-paired
differences rather than with setting fixed effects; with setting fixed effects $p=0.054$ (H4) and $0.17$ (H5), and
no conclusion changes. (iv) A memory failure on the machine that drove the runs stalled one run
(1.5B GSM8K, greedy, $n=256$, seed 13) at generation 6; it was rerun from scratch, and the remaining runs were moved to
drivers on the GPU server itself (same code and model servers). (v) The share of generations with fewer than $K=4$
valid candidates, which the plan asked us to report, is given per arm in Table~\ref{tab:v2arms}. (vi) Secondary
analyses promised in the plan: the Spearman correlation between $n$ and held-out gain of greedy loops is 0.62
($p=0.001$) on 1.5B TREC, 0.74 ($p<0.001$) on 7B TREC and 0.17 ($p=0.42$) on GSM8K; 95\% intervals for the pilot gate
minus McNemar at $n=16$ are $[-5.1,17.3]$, $[-0.7,1.7]$, $[3.8,15.8]$ and $[-8.5,0.0]$ points for 1.5B TREC, Banking77,
7B TREC and GSM8K, and at $n=64$ $[-1.6,13.0]$, $[-4.2,1.1]$, $[-0.2,4.6]$ and $[-10.9,1.7]$. (vii) Sensitivity
figures, computed after the study from the observed variances: the pooled H4 test could have detected an improvement
of 3.2 points with 80\% power (per setting 2.4 to 9.9 points; H1 2.3 to 8.5; H2 8.9 to 15.4).

\paragraph{Current models and GEPA.} GEPA runs use DSPy 3.4.0 (\texttt{dspy.GEPA}, gepa 0.1.4) with default settings
except the budget (4000 metric calls), \texttt{reflection\_minibatch\_size}=3 and \texttt{track\_stats}. The program
is a single \texttt{Predict} module (TREC, Banking77; the label list is a fixed input field) or a
\texttt{ChainOfThought} module (GSM8K), whose instruction starts as the one-sentence seed. The task model Qwen3.5-4B
decodes greedily; the reflection model Qwen3.8-27B (FP8 weights) samples at temperature 1; thinking is disabled for
both. The training pool (256 items) and the validation set are drawn from the task pool excluding both gold sets, the
validation set nested across $n_{\mathrm{val}}$ for a seed. Two problems were found and fixed before any GEPA
result was analyzed, and all runs were restarted from scratch after each fix. The default limit of 1024 open files
made language-model calls fail mid-run (DSPy scores a failed call as a wrong answer), and an 8192-token context window
made reflections fail once evolved instructions grew long. The final runs use a 32768-token window for the reflection
model and contain no infrastructure errors. Runs with $n_{\mathrm{val}}=16$
progress slowly (every iteration waits for a reflection), so the primary comparison uses a common budget of 1000
metric calls, reconstructed exactly from each run's saved state; runs with
$n_{\mathrm{val}}=16$ were stopped after 1030 calls, and five GSM8K runs with $n_{\mathrm{val}}=64$ after 2100 to 2900.
The held-out set is scored for the starting program, the returned program, the five other best candidates by
validation score and five drawn at random; regret is therefore a lower bound. Answers that cannot be parsed count as
wrong, as in GEPA's own metric. The plan recorded in our dated research log before the runs (17:57) specified the full
budget and held-out scores for all candidates; the budget and the scoring sample were changed for runtime reasons
before any run with $n_{\mathrm{val}}\le64$ had finished. GEPA's reported gain (validation score of the returned
program minus that of the starting program) was added as a measure after the first eight runs, all with
$n_{\mathrm{val}}=256$ and the full budget, had finished and their results had been displayed (19:37--19:38; the
session transcript shows that the analysis script already computed the reported gain for them), and one of these runs
had been reconstructed at 1000 calls to test the reconstruction script (20:09) before the budget change (20:15). No
result with $n_{\mathrm{val}}\le64$ had been seen at any of these points. The pre-planned inflation measure carries
the same finding as the reported gain. GEPA checks its budget only when it starts an iteration and records a candidate's
discovery count before its full validation pass, so a run stopped at 1000 calls could also contain a candidate with a
count of 1001 to 1006. In the 21 runs with complete records two such candidates exist and neither changes the returned
program. For the runs with $n_{\mathrm{val}}=16$ the saved states were lost when the servers were released, but the
run logs record every evaluation and every accepted candidate's validation score. Rebuilding each pool from
the log alone reproduces the pool, the discovery counts, the validation scores and the returned program of all 45
reconstructions at 1000 calls, including these 15. The same logs give the
reflection calls made within 1000 metric calls, on average 38 to 45 per run (by task) with 16 validation items, 14 to
16 with 64 and 3 to 5 with 256, so the comparison is matched in metric calls, not in reflection compute. Outputs of runs completed before each restart, including four
that had been scored on the held-out set, were kept but not analyzed. The Qwen3.5-4B self-improvement runs use the loop of Section~\ref{sec:exp} unchanged
(same prompts, feedback pool and confirmatory gold set) with Qwen3.5-4B as solver and proposer: greedy at
$n=16,64,256$ with seeds 11--14 on TREC and Banking77, and one oracle-instrumented run per task ($n=256$, $T=15$).
Budget constraints led us to drop a planned arm in which Qwen3.8-27B proposes for the 1.5B solver.

\paragraph{Follow-up studies.} Four studies were added in response to reviews, each planned in the log before its
first run; all used the software of the GEPA study (vLLM 0.30.0, DSPy 3.4.0, gepa 0.1.4) on two H100 instances. (i)
Strong starts for the loop and for GEPA (Appendix~\ref{app:strong}). (ii) GEPA with 16 validation items from the
original seed, seeds 1--5, run towards 3000 calls on TREC and Banking77, and compared with the full-budget runs with 64
and 256 items truncated at the same budget. Truncating a completed run from its candidate records is exact for the same
reason as the reconstruction from saved state, and it reproduces all 17 state-based reconstructions at 1000 calls whose
selected candidate had been held-out-scored, including the held-out vectors; selected candidates without a held-out
score were scored with the same evaluator. (iii) MIPROv2 (\texttt{dspy.MIPROv2},
optuna 5.0.0), instruction-only, with 16 proposed instructions and 24 Bayesian-optimization trials, each scored on the
full validation set, the same models, training pool and validation sets as GEPA, TREC and Banking77,
$n_{\mathrm{val}}\in\{16,256\}$ and seeds 1--5; the held-out set scores the seed program, the returned program and four
other evaluated candidates. (iv) Four more seeds of the strong-start loop, decided
after the first four seeds' results had been seen and reported separately. GEPA runs still running at 04:30 were stopped
and reconstructed at the largest multiple of 250 calls that every run of their group had reached. This rule was fixed
before any held-out score of these runs was seen, although interim validation scores from iteration 2 of one run
had been visible.

\paragraph{Inference nondeterminism.} Batched decoding is not bitwise reproducible. Within a run, every
(instruction, item) pair is scored once and cached, so an artifact has one score vector. Across runs, the seed
instruction's gold vector differs as follows (largest pairwise share of flipped items). In the exploratory runs: 0\% for
llama.cpp and for vLLM on Banking77, at most 1\% for vLLM on TREC, 9.3\% on GSM8K. In the confirmatory runs, which ran
under heavier and more varied server load: 0.7\% (1.5B TREC), 1.2\% (Banking77), 1.5\% (7B TREC) and 9.8\% on GSM8K,
where the seed's held-out accuracy ranges from 61.0 to 64.3\% across runs. Different GEPA servers likewise give
slightly different scores. Long generations are the most sensitive to batch composition. For GSM8K this item-level
noise is a substantial part of $v$ and of the ``true'' effects, and it is the likely reason GSM8K is among the least
consistent settings in Figure~\ref{fig:breeder}. The pre-registered GSM8K comparisons are insensitive to it. Measuring
every run against the mean seed accuracy instead of its own changes the pilot-versus-greedy difference from $-5.75$ to
$-5.71$ points ($p=0.008$ either way). We also re-scored the seed and all 59 distinct final instructions of the 72
GSM8K runs three times, each time in one batch and bypassing the response cache, on an H100 rather than the study's A100;
the seed scored
62.3, 63.7 and 63.3\%, a final instruction's score varied by 0.7 points across repetitions, and a run's gain moved by
0.7 points on average. With the averaged re-scored gains, H1 on GSM8K is $+1.8\pm2.0$ points ($p=0.40$; originally
$+1.6$, $p=0.48$) and the pilot-versus-greedy difference (H4) is $-5.6\pm1.7$ ($p=0.013$; originally $-5.75$,
$p=0.008$).

\paragraph{Data and licenses.}
TREC \citep{li2002trec} (5{,}952 items, 50 fine labels; \texttt{CogComp/trec}), Banking77
\citep{casanueva2020banking77} (13{,}069 items; CC-BY-4.0; \texttt{PolyAI/banking77}), GSM8K \citep{cobbe2021gsm8k}
(8{,}792 items; MIT). The Qwen2.5, Qwen3.5 and Qwen3.8 models are released under the Qwen or Apache-2.0 licenses.
Compute: one consumer RTX 3080 Ti (llama.cpp replication) and about 28 rented GPU-hours (two A100-80GB and four
H100-80GB instances, vLLM 0.30), of which about 11 for the follow-up studies.

\section{Additional results}
\label{app:more}

\paragraph{Proposal distributions.}
Table~\ref{tab:proposals} summarizes the held-out effects of all candidates of the oracle-instrumented runs, including
those of the 14B proposer (confirmatory study; effects on the confirmatory gold set).

\begin{table}[htbp]
\centering
\caption{\textbf{Proposal distributions} of oracle-instrumented runs: number of candidates; share with a positive held-out effect
over their incumbent and mean held-out effect $\mu$ (points), for the first rewrite of the seed and for later generations;
within-generation spread $s$ (corrected for gold-set noise); $v_{cc}/v$; and excess kurtosis of within-generation
deviations.}
\label{tab:proposals}
\small
\setlength{\tabcolsep}{3pt}
\begin{tabular}{llcccccccc}
\toprule
 & & & \multicolumn{2}{c}{generation 1} & \multicolumn{2}{c}{later generations} & & & \\
Proposer & Solver, task & cand. & better & $\mu$ & better & $\mu$ & $s$ & $v_{cc}/v$ & exc.\ kurt. \\
\midrule
1.5B & 1.5B TREC & 64 & 100\% & +6.8 & 8\% & -13.5 & 7.1 & 0.44 & -0.4 \\
1.5B & 1.5B Banking77 & 77 & 0\% & -11.5 & 4\% & -10.0 & 8.2 & 0.60 & -0.2 \\
1.5B & 1.5B GSM8K & 55 & 0\% & -7.7 & 39\% & -2.3 & 7.5 & 0.52 & 3.4 \\
7B & 7B TREC & 120 & 100\% & +4.9 & 15\% & -7.2 & 7.1 & 0.58 & 2.7 \\
7B & 7B Banking77 & 120 & 25\% & -0.3 & 38\% & -0.2 & 0.6 & 0.57 & 2.5 \\
1.5B & 1.5B TREC (llama.cpp) & 133 & 75\% & +5.0 & 20\% & -6.5 & 7.2 & 0.55 & 0.2 \\
1.5B & 1.5B Banking77 (llama.cpp) & 152 & 25\% & -4.5 & 26\% & -1.9 & 2.7 & 0.56 & 3.2 \\
14B & 1.5B TREC & 60 & 75\% & +10.0 & 9\% & -11.0 & 9.5 & 0.57 & -0.1 \\
14B & 1.5B Banking77 & 60 & 25\% & -6.7 & 20\% & -1.4 & 2.5 & 0.55 & 7.6 \\
14B & 1.5B GSM8K & 60 & 50\% & -4.1 & 20\% & -4.0 & 5.6 & 0.52 & 0.6 \\
Qwen3.5-4B & Qwen3.5-4B TREC & 60 & 100\% & +25.9 & 7\% & -9.8 & 7.0 & 0.44 & 0.5 \\
Qwen3.5-4B & Qwen3.5-4B Banking77 & 60 & 100\% & +14.0 & 20\% & -1.0 & 0.9 & 0.48 & 0.1 \\
\bottomrule
\end{tabular}

\end{table}

\paragraph{Where the gains come from.}
With a one-sentence seed, much of a loop's final gain arrives with its first commit. For greedy loops, the gain after
generation 1 as a share of the final gain is 38 to 54\% for the Qwen2.5 models on TREC (so later generations still
doubled the gain), 70 to 78\% on TREC when Qwen2.5-14B proposes, and 90 to 112\% for Qwen3.5-4B, whose later commits added
nothing on average (Banking77 and GSM8K with Qwen2.5 have final gains too small for a
ratio). The strong-start runs of Section~\ref{sec:gepa} remove the first-rewrite phase by construction.

\paragraph{Confirmatory runs, all arms.}
Table~\ref{tab:v2arms} lists every arm of the pre-registered study: final held-out gain over the seed instruction and
final proxy--held-out gap (mean $\pm$ s.e. over seeds), harmful commits (held-out effect $<0$, and $\le-1.8$ points,
one gold-set standard error) out of all commits, summed over seeds, and the share of generations with fewer than
$K=4$ valid candidates.

\begin{table}[htbp]
\centering
\caption{Confirmatory (v2) runs: all arms.}
\label{tab:v2arms}
\small
\resizebox{\textwidth}{!}{\begin{tabular}{llcccccc}
\toprule
Setting & Rule ($n$) & runs & held-out gain & proxy$-$held-out & harmful / commits & $\le-1.8$ pts & $<K$ cand. \\
\midrule
1.5B TREC & Greedy (16) & 8 & +13.7$\pm$2.9 & +17.3$\pm$2.5 & 4 / 22 & 3 & 19\% \\
 & Greedy (64) & 8 & +18.5$\pm$1.5 & +8.0$\pm$1.2 & 8 / 35 & 5 & 20\% \\
 & Greedy (256) & 8 & +24.2$\pm$0.7 & +1.4$\pm$0.8 & 8 / 46 & 4 & 12\% \\
 & McNemar (16) & 8 & +6.3$\pm$4.1 & +5.9$\pm$2.4 & 0 / 2 & 0 & 2\% \\
 & McNemar (64) & 8 & +14.6$\pm$2.7 & +6.2$\pm$2.5 & 2 / 12 & 1 & 9\% \\
 & Select-confirm (16) & 8 & +15.8$\pm$3.2 & +11.3$\pm$4.8 & 0 / 10 & 0 & 13\% \\
 & Select-confirm (64) & 8 & +17.7$\pm$1.9 & +7.0$\pm$1.3 & 3 / 23 & 2 & 13\% \\
 & Pilot gate (16) & 8 & +12.4$\pm$2.9 & +13.9$\pm$3.1 & 1 / 12 & 1 & 11\% \\
 & Pilot gate (64) & 8 & +20.2$\pm$1.0 & +5.5$\pm$1.9 & 4 / 22 & 2 & 9\% \\
\midrule
1.5B Banking77 & Greedy (16) & 8 & -0.6$\pm$0.7 & +13.2$\pm$3.7 & 5 / 12 & 4 & 20\% \\
 & Greedy (64) & 8 & +0.0$\pm$0.5 & +5.3$\pm$1.9 & 6 / 15 & 2 & 10\% \\
 & Greedy (256) & 8 & -0.5$\pm$0.5 & +2.7$\pm$1.1 & 12 / 21 & 2 & 4\% \\
 & McNemar (16) & 8 & +0.0$\pm$0.0 & +0.0$\pm$4.8 & 0 / 0 & 0 & 1\% \\
 & McNemar (64) & 8 & -0.2$\pm$0.2 & +1.4$\pm$2.0 & 1 / 1 & 0 & 4\% \\
 & Select-confirm (16) & 8 & -2.3$\pm$1.7 & +10.2$\pm$4.8 & 3 / 5 & 3 & 2\% \\
 & Select-confirm (64) & 8 & -1.1$\pm$0.9 & +3.9$\pm$2.3 & 4 / 4 & 1 & 4\% \\
 & Pilot gate (16) & 8 & +0.5$\pm$0.5 & +10.5$\pm$4.7 & 2 / 8 & 2 & 4\% \\
 & Pilot gate (64) & 8 & -1.7$\pm$1.1 & +7.2$\pm$2.0 & 7 / 10 & 2 & 8\% \\
\midrule
7B TREC & Greedy (16) & 8 & +14.0$\pm$0.7 & +16.6$\pm$4.4 & 3 / 19 & 1 & 0\% \\
 & Greedy (64) & 8 & +19.8$\pm$0.5 & +5.4$\pm$1.1 & 5 / 34 & 0 & 0\% \\
 & Greedy (256) & 8 & +20.4$\pm$0.6 & +1.3$\pm$0.8 & 16 / 54 & 3 & 0\% \\
 & McNemar (16) & 8 & +1.8$\pm$1.8 & +8.6$\pm$4.5 & 0 / 1 & 0 & 0\% \\
 & McNemar (64) & 8 & +13.6$\pm$1.3 & +1.9$\pm$1.4 & 0 / 8 & 0 & 0\% \\
 & Select-confirm (16) & 8 & +7.6$\pm$1.9 & +17.5$\pm$3.7 & 1 / 10 & 0 & 0\% \\
 & Select-confirm (64) & 8 & +15.9$\pm$0.9 & +4.0$\pm$1.2 & 3 / 21 & 0 & 0\% \\
 & Pilot gate (16) & 8 & +11.5$\pm$2.4 & +16.8$\pm$3.5 & 3 / 15 & 3 & 0\% \\
 & Pilot gate (64) & 8 & +15.8$\pm$1.7 & +5.8$\pm$0.9 & 1 / 23 & 1 & 0\% \\
\midrule
1.5B GSM8K & Greedy (16) & 8 & +1.5$\pm$1.7 & +19.6$\pm$4.5 & 7 / 16 & 5 & 8\% \\
 & Greedy (64) & 8 & -1.1$\pm$1.1 & +12.2$\pm$2.2 & 15 / 27 & 12 & 9\% \\
 & Greedy (256) & 8 & +3.1$\pm$1.0 & +4.9$\pm$1.0 & 13 / 29 & 8 & 12\% \\
 & McNemar (16) & 8 & +0.0$\pm$0.0 & +1.7$\pm$3.5 & 0 / 0 & 0 & 1\% \\
 & McNemar (64) & 8 & +2.7$\pm$1.8 & +7.7$\pm$2.9 & 2 / 7 & 1 & 8\% \\
 & Select-confirm (16) & 8 & +1.1$\pm$1.3 & +13.2$\pm$5.1 & 1 / 6 & 1 & 5\% \\
 & Select-confirm (64) & 8 & +1.9$\pm$2.0 & +8.8$\pm$2.9 & 6 / 12 & 3 & 6\% \\
 & Pilot gate (16) & 8 & -4.3$\pm$1.8 & +22.2$\pm$4.9 & 9 / 15 & 8 & 6\% \\
 & Pilot gate (64) & 8 & -1.9$\pm$1.3 & +12.6$\pm$2.8 & 13 / 27 & 12 & 15\% \\
\midrule
14B$\to$1.5B TREC & Greedy (16) & 5 & +26.1$\pm$1.5 & +22.3$\pm$6.5 & 4 / 15 & 3 & 0\% \\
 & Greedy (64) & 5 & +32.2$\pm$1.7 & +8.1$\pm$2.4 & 5 / 24 & 4 & 0\% \\
 & Greedy (256) & 5 & +32.8$\pm$2.0 & -0.8$\pm$1.5 & 5 / 24 & 1 & 0\% \\
\midrule
14B$\to$1.5B Banking77 & Greedy (16) & 5 & -2.3$\pm$1.1 & +13.7$\pm$1.3 & 6 / 9 & 4 & 0\% \\
 & Greedy (64) & 5 & -1.3$\pm$0.9 & +9.0$\pm$1.0 & 9 / 14 & 3 & 0\% \\
 & Greedy (256) & 5 & +1.6$\pm$0.5 & +3.2$\pm$1.0 & 5 / 16 & 2 & 0\% \\
\midrule
14B$\to$1.5B GSM8K & Greedy (16) & 5 & -3.3$\pm$3.5 & +21.0$\pm$5.1 & 5 / 8 & 4 & 0\% \\
 & Greedy (64) & 5 & +2.4$\pm$2.2 & +10.6$\pm$2.7 & 3 / 9 & 2 & 0\% \\
 & Greedy (256) & 5 & +3.6$\pm$1.0 & +4.3$\pm$0.6 & 2 / 13 & 1 & 0\% \\
\bottomrule
\end{tabular}
}
\end{table}

\paragraph{Incumbent inflation by generation and selection-set size.}
Table~\ref{tab:lockin} gives the incumbent inflation of Section~\ref{sec:lockin}. In the vLLM oracle runs, the share
of candidates whose measured effect on the reused selection set is positive falls from 61\% in the first generation
to 1.7\% in generations 11 to 20. Whether this decline is genuine depends on the threshold. The share with a positive
held-out effect falls only from 50\% to 16\%, while the share that improves by more than one gold-set standard
error (1.8 points) falls from 32\% to 2.3\%, about as fast as the measured share. The 14\% of late candidates with
held-out effects between 0 and 1.8 points lie within one gold-set standard error of zero. If they are real gains,
greedy acceptance misses them, consistently with lock-in, but they are also consistent with no gain.
42\% of all commits occur in the first three generations.

\begin{table}[htbp]
\centering
\caption{\textbf{Incumbent inflation.} Incumbent's selection-set minus held-out score, relative to the seed's offset (points;
greedy runs, all tasks and models; exploratory vLLM runs and pre-registered confirmatory runs; generations 11 to 20:
mean $\pm$ s.e.\ over runs). The exploratory rows for $n=32$ and $128$ contain fewer settings (no 7B Banking77).}
\label{tab:lockin}
\small
\begin{tabular}{llccc}
\toprule
Study & $n$ & gens 1--3 & gens 4--10 & gens 11--20 \\
\midrule
exploratory & 16 & 9.4 & 13.3 & 16.5$\pm$2.6 \\
 & 32 & 6.2 & 8.6 & 10.2$\pm$2.8 \\
 & 64 & 3.7 & 5.3 & 7.3$\pm$1.7 \\
 & 128 & 2.3 & 3.0 & 4.2$\pm$1.4 \\
 & 256 & 1.6 & 2.1 & 2.5$\pm$0.7 \\
\midrule
confirmatory & 16 & 7.6 & 11.8 & 13.6$\pm$1.7 \\
 & 64 & 3.6 & 5.9 & 7.2$\pm$1.1 \\
 & 256 & 0.0 & 0.7 & 1.7$\pm$0.6 \\
\bottomrule
\end{tabular}

\end{table}

\paragraph{Commits in native loops.}
Tables~\ref{tab:commits} and~\ref{tab:commits_setting} give the comparison of Section~\ref{sec:lockin}, a post hoc,
descriptive analysis of every commit of the 96 greedy runs of the confirmatory study. For a commit in generation $t$,
the prior $(\mu,s^2)$ is the frozen phase-indexed pilot prior of the same model and task (generation 1 or later;
Appendix~\ref{app:alloc}), fitted to held-out effects of the separate exploratory oracle runs before these runs
started. The noise terms are $\hat e_\eta^2=\hat v_{cc}/n$ and $\hat e_c^2=\max(\hat v-\hat v_{cc},0)/n$, computed
from the item vectors of that generation's candidates on the loop's selection set (for a generation with a single
valid candidate, $\hat v_{cc}=\hat v/2$). We simulate Assumption~\ref{ass:gauss} with the
generation's number of valid candidates (40{,}000 draws) and average the overstatement of the best-measured
candidate over the draws in which its measured gain is positive. The observed overstatement is the commit's measured
gain minus the change in the incumbent's held-out accuracy. Since the same quantity for the incumbent's selection-set
score gives its inflation, a commit's observed overstatement is exactly the change in the incumbent's inflation.
Commits fall into three groups. Generation-1 commits are the only ones Assumption~\ref{ass:gauss} describes. Delayed
first commits also have the seed as incumbent, but they follow one or more rejections on the same items: if the
incumbent's shared error $C$ persists, first acceptance at generation $t$ reweights its distribution by
$(1-p(C))^{t-1}p(C)$ rather than $p(C)$, where $p(C)$ is the acceptance probability given $C$. These commits also
receive the later-generation prior, although their incumbent is still the seed. Later commits are made against an
incumbent selected on the same items. For the last two groups the model value is a fresh-set reference, not a
prediction. The analysis uses no prior fitted to held-out effects from these runs; its noise estimates come from the
current generation's selection-set scores. Intervals bootstrap runs (4{,}000 resamples) and are conditional on the
pilot prior and on the fixed 600 gold items, so they do not reflect uncertainty about a new pilot or a new population
of evaluation items. An interval that contains zero shows that the average residual is compatible with zero, not that
the model is calibrated.

\begin{table}[htbp]
\centering
\caption{\textbf{Overstatement of commits in native greedy loops} (points; confirmatory runs, four settings, eight
seeds, pooled over settings). Observed: measured gain on the selection set minus held-out effect, averaged over
commits. Model: Assumption~\ref{ass:gauss} with the frozen pilot prior and the generation's own noise terms,
conditioned on acceptance or not. Generation 1: commits in the first generation; delayed first: a run's first commit
made after one or more rejections; later: commits made against an incumbent selected on the same items. Runs: runs
contributing commits. 95\% intervals bootstrap runs and are conditional on the pilot and the gold items.}
\label{tab:commits}
\small
\resizebox{\textwidth}{!}{\begin{tabular}{llcccccc}
\toprule
$n$ & commits & number & runs & observed & model, accepted & model, unconditional & obs.\ $-$ model (95\% CI) \\
\midrule
16 & generation 1 & 21 & 21 & 9.1 & 9.0 & 6.0 & +0.0 [-3.4, +4.0] \\
 & delayed first & 10 & 10 & 6.3 & 9.6 & 6.2 & -3.3 [-8.0, +1.8] \\
 & later & 38 & 22 & 5.2 & 9.7 & 5.7 & -4.5 [-6.1, -2.6] \\
\midrule
64 & generation 1 & 27 & 27 & 3.2 & 3.8 & 2.2 & -0.5 [-2.5, +1.5] \\
 & delayed first & 4 & 4 & 3.5 & 3.8 & 2.1 & -0.2 [-1.9, +1.3] \\
 & later & 80 & 28 & 1.9 & 3.7 & 1.9 & -1.8 [-2.5, -1.1] \\
\midrule
256 & generation 1 & 24 & 24 & 0.1 & 1.3 & 0.7 & -1.2 [-2.7, +0.4] \\
 & delayed first & 8 & 8 & 0.6 & 1.3 & 0.7 & -0.6 [-3.1, +1.3] \\
 & later & 118 & 31 & 0.5 & 1.1 & 0.5 & -0.6 [-1.0, -0.2] \\
\bottomrule
\end{tabular}
}
\end{table}

\begin{table}[htbp]
\centering
\caption{\textbf{Commits in native greedy loops by setting} (points; observed / model value conditioned on
acceptance, with the number of commits in parentheses; groups as in Table~\ref{tab:commits}).}
\label{tab:commits_setting}
\small
\begin{tabular}{llccc}
\toprule
Setting & $n$ & generation 1 & delayed first & later \\
\midrule
1.5B TREC & 16 & 1.9 / 5.9 (6) & 9.3 / 10.3 (2) & 5.7 / 11.8 (14) \\
 & 64 & -0.4 / 1.5 (7) & 1.1 / 3.6 (1) & 1.9 / 4.4 (27) \\
 & 256 & -0.5 / 0.5 (8) & -- & -0.0 / 1.4 (38) \\
\midrule
1.5B Banking77 & 16 & 12.9 / 7.0 (4) & 11.1 / 9.6 (3) & 4.1 / 7.5 (5) \\
 & 64 & 6.6 / 3.2 (5) & 3.4 / 3.6 (2) & 1.2 / 2.5 (8) \\
 & 256 & 0.4 / 1.0 (3) & 1.0 / 1.3 (5) & 1.4 / 1.0 (13) \\
\midrule
7B TREC & 16 & 9.6 / 7.6 (6) & -0.9 / 9.1 (2) & 3.4 / 7.9 (11) \\
 & 64 & 1.1 / 2.7 (8) & -- & 1.0 / 3.5 (26) \\
 & 256 & -2.1 / 0.8 (7) & -7.1 / 1.3 (1) & 0.5 / 1.0 (46) \\
\midrule
1.5B GSM8K & 16 & 13.9 / 16.1 (5) & 4.4 / 9.5 (3) & 7.7 / 10.1 (8) \\
 & 64 & 6.8 / 7.7 (7) & 6.2 / 4.4 (1) & 3.4 / 3.6 (19) \\
 & 256 & 3.3 / 3.2 (6) & 3.6 / 1.2 (2) & 0.9 / 1.0 (21) \\
\bottomrule
\end{tabular}

\end{table}

\paragraph{Estimating the winner's curse and the gain within a loop.}\label{app:online}
The causal estimate of Section~\ref{sec:measure} uses, for each oracle run, a random subset $A$ of $n$ items of the
run's own selection set $D$ (on which its incumbents were chosen), held fixed for the whole run. At generation $t$ it
computes $\hat v_{cc}$ and the within-generation spread of the measured effects on $A$ from generations 1 to $t$ of
that run only, weighting both by $K_t-1$ for each generation with $K_t\ge2$ valid candidates. It sets
$\hat s^2=\max(\text{spread}-\hat v_{cc}/n,0)$ and $\hat h_t=\hat s^2/(\hat s^2+\hat v_{cc}/n)$,
and predicts the true advantage of the generation's winner over the generation mean as $\hat h_t$ times its measured
advantage. The truth is the advantage on the 600 gold items (100 draws of $A$ per run). All errors are of averages over a setting's generations, runs and draws of $A$; the prediction for
a single winner is much less precise. Over nine settings and $n\in\{16,32,64,128\}$ the predicted ratio of true to
measured advantage has a mean absolute error of 0.10 (0.08 at $n=16$) and a bias of $+0.02$; restricted to the first
three generations, when the loop has almost no history, the error is 0.13 and the bias $-0.01$. The estimate is too
high where candidates barely differ (7B and Qwen3.5-4B on Banking77, where the floor at zero keeps $\hat s^2$ positive)
and on GSM8K, and too low at $n=128$ on the 1.5B vLLM settings, where $A$ is half of the reused set. Because the
oracle runs select on 256 items, $A$ is a subset of a larger reused set and the trajectory is not that of a loop with
$n$ items; Table~\ref{tab:commits} is the check on native loops.

Table~\ref{tab:online} asks how a loop should report its gain. For every non-oracle run of the confirmatory study, the
14B-proposer runs and the Qwen3.5-4B runs, we take the incumbent entering the last generation and compare three
estimates of its gain over the seed with the gain on a random half of the gold set: the loop's self-report on its
selection set; the same sum over commits with each committed candidate's measured advantage over its generation mean
shrunk by $\hat h_t$ (no extra evaluations); and the gain on $m$ audit items from the other half of the gold set, which
are scored only for the seed and the incumbent after 19 generations and never used for selection (50 random halvings). Shrinkage largely removes the average bias but not the setting-level bias ($-8.5$ to $+5.8$
points at $n=16$ with greedy acceptance). On TREC, where the first commits are large real gains, it removes too much,
and on Banking77 and GSM8K it removes too little, because the shared error that the incumbent accumulates on a reused
set moves the measured effects of all candidates together. The audit is unbiased by construction, and 64 items scored for these two instructions (128 evaluations) already
bring the error of the reported gain at $n=16$ below that of a self-report with $n=64$.

\begin{table}[htbp]
\centering
\caption{\textbf{Reporting the gain of a loop.} Bias / root-mean-square error (points) of three estimates of the
incumbent's gain over the seed after 19 generations, against the gain on 300 held-out gold items, by selection-set
size $n$ (all rules). Extra evals are additional to the nominal candidate-selection budget $KnT=80n$,
which excludes seed and feedback scoring; candidate shortfalls reduce the actual count.}
\label{tab:online}
\small
\setlength{\tabcolsep}{3pt}
\begin{tabular}{lcccc}
\toprule
Estimate of the gain & extra evals & $n=16$ & $n=64$ & $n=256$ \\
\midrule
self-report (selection set) & 0 & +10.8 / 15.5 & +6.1 / 9.0 & +1.8 / 4.7 \\
shrinkage by the loop's own $\hat h_w^2$ & 0 & +0.7 / 10.7 & -0.7 / 7.6 & -1.4 / 5.3 \\
audit, 16 items & 32 & -0.0 / 10.7 & -0.3 / 11.9 & +0.1 / 12.3 \\
audit, 32 items & 64 & -0.1 / 7.7 & -0.2 / 8.7 & -0.0 / 9.0 \\
audit, 64 items & 128 & -0.0 / 5.7 & -0.1 / 6.3 & -0.0 / 6.7 \\
audit, 128 items & 256 & -0.1 / 4.4 & -0.1 / 4.9 & +0.0 / 5.1 \\
\midrule
runs & & 151 & 151 & 55 \\
mean true gain & & 6.2 & 9.6 & 14.2 \\
\bottomrule
\end{tabular}

\end{table}

\paragraph{Strong starts.}\label{app:strong}
Both follow-up studies were planned in our dated research log before their first run (23:38--23:39). The loop starts
from the final instruction of the Qwen3.5-4B greedy run with $n=256$ and seed 11, fixed a priori rather than chosen by
held-out accuracy, which would bias later gains downward by regression to the mean (held-out accuracy 63.5\% on TREC
and 65.0\% on Banking77). The a-priori choice happened to be the weakest of the four candidate runs on TREC (the others
reached 66.0 to 68.5\%) and the second weakest on Banking77 (64.5 to 68.8\%), so the absolute gains of both arms may
partly reflect regression to the mean in the other direction; the comparison between arms is unaffected. It runs greedy acceptance with $K=4$, $T=20$, $n\in\{16,256\}$ and seeds 21--24, plus one
oracle-instrumented run per task ($n=256$, $T=10$). Four further seeds (25--28) were added after the results of the
first four had been seen. The planned seeds alone gave held-out differences (256 minus 16 items) of $6.5\pm5.3$ points
on TREC and $1.8\pm1.4$ on Banking77 ($p=0.31$ each) and falls in the offset-corrected overstatement of 13.6 and 12.5
points ($p=0.095$ and $<0.001$); the planned proxy--held-out gap, which carries the start's offset on each 16-item
set, fell by 9.4 and 2.8 points ($p=0.12$ and $0.69$). The four added seeds gave similar results, with held-out
differences of 5.6 and 1.9 points ($p=0.15$ and $0.16$), gap reductions of 7.2 and 12.9 points ($p=0.08$ and 0.03)
and overstatement falls of 16.2 and 7.5 points ($p=0.04$ and $0.09$). Section~\ref{sec:gepa} also reports an
inverse-normal combination of the two stages with equal weights (one-sided stage-wise $p$-values). The decision to
run the second stage was made after the first had been seen, and the weights and the one-sided direction were chosen
after both stages had run, so the combined $p$-values are exploratory. The table pools all eight seeds. GEPA starts from the program returned by its full-budget run with
256 validation items and seed 1 (held-out accuracy 69.5\% on TREC and 64.0\% on Banking77; instructions of 755 and
2151 words), with seeds 6--10, $n_{\mathrm{val}}\in\{16,256\}$ and a budget of 3000 calls, cut by the time rule of
Appendix~\ref{app:alloc} to 2750 calls on TREC and 2000 on Banking77. With 16 validation items GEPA's pool held 114
and 80 programs, with 256 items 11 and 8; the returned program's validation score implied gains of 25.0 points on
both tasks with 16 items, against real gains of $1.6\pm0.6$ and $4.1\pm0.7$, and 3.0 and 3.7 points with 256 items,
against $0.8\pm0.8$ and $1.9\pm1.1$. The offset-corrected overstatement fell by 21.2 and 19.2 points ($p=0.003$ and
0.03), while the held-out gains did not differ significantly ($p=0.51$ and 0.13). Reconstructing with the six-call
window of GEPA's stopping rule changes no conclusion (no run's measures move by more than 0.2 points; for the
matched-budget runs of Table~\ref{tab:gepafollow} nothing changes).
Table~\ref{tab:strongstart} gives all measures.

\begin{table}[htbp]
\centering
\caption{\textbf{Strong starts} (points; mean $\pm$ s.e.; eight seeds for the loops, five for GEPA). Reported: gain
over the start on the selection (validation) set; held-out: gain on the 600 gold items; gap / inflation: final
selection-set minus held-out accuracy.}
\label{tab:strongstart}
\small
\begin{tabular}{llcccccc}
\toprule
 & & \multicolumn{3}{c}{GEPA, $B^\star$ metric calls} & \multicolumn{3}{c}{greedy loop, 20 generations} \\
\cmidrule(lr){3-5}\cmidrule(lr){6-8}
Task & $n$ & reported & held-out & inflation & reported & held-out & gap \\
\midrule
TREC & 16 & +25.0$\pm$4.4 & +1.6$\pm$0.6 & +16.4$\pm$3.1 & +14.8$\pm$3.7 & -1.5$\pm$2.7 & +9.1$\pm$2.5 \\
 & 256 & +3.0$\pm$1.1 & +0.8$\pm$0.8 & -0.9$\pm$0.8 & +6.0$\pm$1.0 & +4.6$\pm$0.9 & +0.7$\pm$1.0 \\
\midrule
Banking77 & 16 & +25.0$\pm$5.2 & +4.1$\pm$0.7 & +22.1$\pm$2.1 & +11.7$\pm$2.2 & +0.2$\pm$0.4 & +9.0$\pm$3.6 \\
 & 256 & +3.7$\pm$0.6 & +1.9$\pm$1.1 & +3.1$\pm$1.3 & +3.6$\pm$0.7 & +2.0$\pm$0.5 & +1.2$\pm$0.6 \\
\bottomrule
\end{tabular}

\end{table}

\paragraph{GEPA, all conditions.}
Table~\ref{tab:gepa} gives the GEPA results at the common budget of 1000 metric calls, and Table~\ref{tab:gepafull}
those of the runs that completed the full budget of 4000 calls. The full budget shows the same pattern. The
returned program's inflation is larger with 64 than with 256 validation items (TREC $4.5$ against $0.0$ points,
Banking77 $12.7$ against $5.2$), and so is the overstatement of its gain (7.2 against 1.4 and 10.5 against 4.6
points), while its held-out gain is not smaller. Table~\ref{tab:gepafollow} adds the
follow-up runs. With 16 validation items run to 2250 calls from the one-line seed (seeds 1--5, compared with the
full-budget 64- and 256-item runs truncated at 2250 calls), the pool held 91 programs against 32 and 9, and the 16-item
arm made on average 107 reflection calls against 10 or fewer with 256 items. Held-out gains did not grow with
$n_{\mathrm{val}}$ (256 minus 16 items: $+0.5\pm2.5$ on TREC, $-2.8\pm1.6$ on Banking77; $p=0.86$ and 0.15), while
the offset-corrected overstatement fell by 28.4 and 15.1 points ($p=0.007$ and 0.002). Reconstructed at 1000 calls,
the new 16-item runs replicate the original arm (inflation 19.8 against 14.1 points on TREC, 23.1 against 24.6 on
Banking77; held-out gain 17.5 against 19.2 and 2.9 against 1.2). Every reconstruction of the follow-up runs agrees
with the run's log in pool, discovery counts, validation scores and returned program. The 16-item arm at 2250 calls consists of new runs on different servers from those of the truncated 64-
and 256-item runs (same seeds, training pools and nested validation sets); the difference between servers moves
held-out scores by about a point (Appendix~\ref{app:alloc}). Proposition~\ref{prop:breeder}'s prediction (last column)
now takes $v_{cc}$ from the per-item validation scores that GEPA logs for each accepted candidate, so that it uses
only what the optimizer records. Where the overstatement is large, the observed
value is 1.1 to 1.9 times the prediction across all GEPA conditions (median 1.4; at 1000 calls 1.1 to 1.4, for the
16-item arms at the larger budgets 1.5 to 1.9); with 256 items both are small (observed $-3.0$ to $+3.7$ points,
predicted 0.5 to 1.9). The model thus captures how the overstatement depends on the validation-set size, the pool size and the
budget, but not its level for GEPA, whose pool is itself selected on the validation set.

\begin{table}[htbp]
\centering
\caption{\textbf{GEPA follow-up runs} (points; mean $\pm$ s.e.\ over five seeds). $B$: metric calls at which all runs
of a group are compared. Cand.: programs in GEPA's pool. Overstatement: reported minus held-out gain. Prop.~1:
predicted overstatement.}
\label{tab:gepafollow}
\small
\resizebox{\textwidth}{!}{\begin{tabular}{lllcccccc}
\toprule
Start & Task & $n_{\mathrm{val}}$ & $B$ & cand. & reported gain & held-out gain & overstatement & Prop.~1 \\
\midrule
one-line seed & TREC & 16 & 2250 & 91 & +45.0$\pm$7.2 & +18.5$\pm$1.2 & +26.5$\pm$6.9 & 14.1 \\
 &  & 64 &  & 32 & +26.6$\pm$4.0 & +20.4$\pm$1.6 & +6.2$\pm$3.9 & 5.4 \\
 &  & 256 &  & 9 & +17.0$\pm$2.6 & +19.0$\pm$1.9 & -1.9$\pm$2.2 & 0.8 \\
 & Banking77 & 16 & 2250 & 91 & +22.5$\pm$3.2 & +3.7$\pm$1.4 & +18.8$\pm$2.5 & 12.6 \\
 &  & 64 &  & 32 & +11.2$\pm$2.3 & +1.3$\pm$1.1 & +10.0$\pm$2.1 & 5.5 \\
 &  & 256 &  & 9 & +4.5$\pm$1.0 & +0.8$\pm$0.6 & +3.7$\pm$0.7 & 1.8 \\
\midrule
strong start & TREC & 16 & 2750 & 114 & +25.0$\pm$4.4 & +1.6$\pm$0.6 & +23.4$\pm$4.1 & 15.2 \\
 &  & 256 &  & 11 & +3.0$\pm$1.1 & +0.8$\pm$0.8 & +2.2$\pm$1.2 & 1.8 \\
 & Banking77 & 16 & 2000 & 80 & +25.0$\pm$5.2 & +4.1$\pm$0.7 & +20.9$\pm$5.1 & 12.2 \\
 &  & 256 &  & 8 & +3.7$\pm$0.6 & +1.9$\pm$1.1 & +1.7$\pm$1.6 & 1.7 \\
\bottomrule
\end{tabular}
}
\end{table}

\begin{table}[htbp]
\centering
\caption{\textbf{GEPA at a common budget of 1000 metric calls} (points; mean $\pm$ s.e.\ over seeds). Inflation:
validation minus held-out accuracy of the returned program. Regret: best held-out accuracy among scored candidates
minus that of the returned program; it is a lower bound computed over the seed, the returned program and ten others
(5 best, 5 random), which is the whole pool at $n_{\mathrm{val}}=256$ but a quarter of it at $n_{\mathrm{val}}=16$, so
it is not comparable across $n_{\mathrm{val}}$.}
\label{tab:gepa}
\small
\setlength{\tabcolsep}{3pt}
\begin{tabular}{llccccc}
\toprule
Task & $n_{\mathrm{val}}$ & runs & inflation & reported gain & held-out gain & regret \\
\midrule
TREC & 16 & 5 & +14.1$\pm$2.7 & +38.8$\pm$4.6 & +19.2$\pm$2.0 & +4.9$\pm$1.9 \\
 & 64 & 5 & +2.4$\pm$2.6 & +24.4$\pm$4.3 & +19.3$\pm$1.7 & +2.1$\pm$0.6 \\
 & 256 & 5 & -4.5$\pm$1.3 & +15.9$\pm$2.9 & +18.9$\pm$1.5 & +0.2$\pm$0.1 \\
\midrule
Banking77 & 16 & 5 & +24.6$\pm$5.0 & +18.8$\pm$2.8 & +1.2$\pm$1.7 & +1.8$\pm$0.6 \\
 & 64 & 5 & +7.3$\pm$2.9 & +7.5$\pm$2.2 & +2.4$\pm$1.0 & +0.3$\pm$0.3 \\
 & 256 & 5 & +2.4$\pm$0.6 & +2.9$\pm$0.7 & +1.2$\pm$0.6 & +0.1$\pm$0.1 \\
\midrule
GSM8K & 16 & 5 & +3.6$\pm$1.3 & +6.2$\pm$2.0 & -0.9$\pm$0.3 & +0.9$\pm$0.3 \\
 & 64 & 5 & +0.6$\pm$1.2 & +3.4$\pm$1.2 & -0.8$\pm$0.5 & +0.8$\pm$0.5 \\
 & 256 & 5 & -0.5$\pm$0.6 & +1.0$\pm$0.5 & -0.4$\pm$0.2 & +0.4$\pm$0.2 \\
\bottomrule
\end{tabular}

\end{table}

\begin{table}[htbp]
\centering
\caption{GEPA runs that completed the full budget of 4000 metric calls (same format as Table~\ref{tab:gepa}; regret
over the 12 best and 12 random candidates scored at the full budget).}
\label{tab:gepafull}
\small
\setlength{\tabcolsep}{3pt}
\begin{tabular}{llccccc}
\toprule
Task & $n_{\mathrm{val}}$ & runs & inflation & reported gain & held-out gain & regret \\
\midrule
TREC & 64 & 5 & +4.5$\pm$3.0 & +28.8$\pm$2.8 & +21.6$\pm$1.0 & +1.4$\pm$0.6 \\
 & 256 & 5 & -0.0$\pm$1.1 & +19.5$\pm$2.4 & +18.1$\pm$1.8 & +2.0$\pm$0.8 \\
\midrule
Banking77 & 64 & 5 & +12.7$\pm$1.6 & +13.4$\pm$2.8 & +2.9$\pm$1.3 & +2.3$\pm$0.7 \\
 & 256 & 5 & +5.2$\pm$0.5 & +5.1$\pm$1.0 & +0.5$\pm$0.7 & +1.8$\pm$0.3 \\
\bottomrule
\end{tabular}

\end{table}

\paragraph{MIPROv2.}\label{app:mipro}
Table~\ref{tab:mipro} gives the MIPROv2 results. In every seed the two validation-set sizes drew the same proposed
instructions from the language-model cache (15 of 16 in one Banking77 seed), so they differ mainly in the validation
set on which MIPROv2 evaluates and selects (and therefore in the sequence of Bayesian-optimization trials). On TREC the larger set doubled the held-out gain of the returned program (11.4 against 5.8
points; paired difference $5.6\pm0.7$, $p=0.001$, all five seeds) and removed its overstatement relative to the seed
(0.7 against 5.5 points, $p=0.05$). The raw inflation is uninformative there, because the seed program's own
validation offset ranges from $-30$ to $+7$ points across the 16-item sets. On Banking77, where candidates barely
differ, the 16-item arm returned programs that were $1.6\pm0.9$ points worse than the seed while their validation
scores implied a 5-point gain; the overstatement relative to the seed was 6.6 points with 16 items and 2.4 with 256
($p=0.27$). The raw inflation (13.8 against 3.0 points, $p=0.02$) mostly reflects the seed program's own validation
offsets on the 16-item sets, which range from $-8$ to $+17$ points. Three of the ten 16-item runs returned the seed
program. In these runs six to nine candidates tied for the best validation score on 16 items, and MIPROv2 kept the
first.
Proposition~\ref{prop:breeder}, with $v_{cc}$ from the candidates' validation item vectors, predicts 9.0, 0.9, 7.7 and
1.9 points for the four cells against observed 5.5, 0.7, 6.6 and 2.4; it overpredicts the 16-item TREC cell, whose
ties the Gaussian model does not describe.

\begin{table}[htbp]
\centering
\caption{\textbf{MIPROv2} (points; mean $\pm$ s.e.\ over five seeds). Cand.: distinct evaluated instructions.
Overstatement: reported minus held-out gain. Prop.~1: predicted overstatement.}
\label{tab:mipro}
\small
\setlength{\tabcolsep}{2pt}
\begin{tabular}{llcccccc}
\toprule
Task & $n_{\mathrm{val}}$ & cand. & inflation & reported gain & held-out gain & overstatement & Prop.~1 \\
\midrule
TREC & 16 & 15.8 & +0.2$\pm$6.5 & +11.2$\pm$4.1 & +5.8$\pm$2.1 & +5.5$\pm$2.6 & 9.0 \\
 & 256 & 14.4 & -0.8$\pm$0.6 & +12.1$\pm$3.1 & +11.4$\pm$2.3 & +0.7$\pm$1.3 & 0.9 \\
\midrule
Banking77 & 16 & 15.8 & +13.8$\pm$2.9 & +5.0$\pm$2.3 & -1.6$\pm$0.9 & +6.6$\pm$3.3 & 7.7 \\
 & 256 & 15.4 & +3.0$\pm$0.4 & +2.5$\pm$0.3 & +0.1$\pm$0.5 & +2.4$\pm$0.4 & 1.9 \\
\bottomrule
\end{tabular}

\end{table}

\begin{figure}[htbp]
\centering
\includegraphics[width=\textwidth]{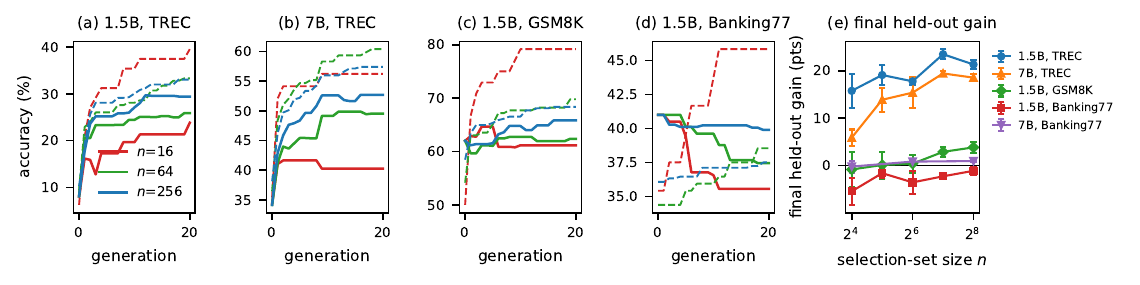}
\caption{\textbf{Exploratory greedy loops (v1, vLLM).} (a--d) Held-out accuracy (solid) and the loop's own
selection-set score (dashed) for $n=16,64,256$, mean of 3 seeds. (e) Final held-out gain over the seed instruction
versus $n$ (mean $\pm$ s.e.).}
\label{fig:main_v1}
\end{figure}

\paragraph{Exploratory greedy loops.}
Figure~\ref{fig:main_v1} shows the exploratory (v1) greedy loops summarized in Section~\ref{sec:exp}.

\paragraph{A hand-written reference instruction.}
To calibrate the size of the TREC gains, we wrote an instruction that defines all 50 labels of the TREC taxonomy
(141 words). On the confirmatory gold set it raises accuracy from 10.5\% to 11.3\%
for the 1.5B model and from 37.2\% to 54.5\% for the 7B model. For 7B the label glossary is worth about as much as the
greedy loops' gain at large $n$; the 1.5B loops gain far more than the glossary (their best final instructions
describe the coarse categories, give worked examples and constrain the output format).

\paragraph{Acceptance decisions in isolation.}\label{sec:decisions}
Proposition~\ref{prop:bayes} concerns a single decision on a fresh selection set, whereas whole runs also involve
adaptive reuse, so we also evaluate the rules one decision at a time (exploratory). At $n=16$ the average greedy decision
lowers held-out accuracy by 0.47 points (95\% interval $[-0.82,-0.03]$), consistent with
Proposition~\ref{prop:goodhart} for $\mu<0$, and every rule with a calibrated threshold avoids the loss. The
pilot-calibrated Bayes gate gains 0.77 points over greedy $[0.60, 0.93]$ and a McNemar test with pilot-tuned $\alpha$
gains 0.68 $[0.43, 0.91]$, and the two are indistinguishable at every $n$. Both avoid the loss mainly by committing
less often, and about half of their commits are still harmful. At $n=16$ greedy commits in 44\% of decisions, the pilot gate in 8\% and the tuned McNemar test in 6\%, and 58\%, 59\% and
46\% of their commits are harmful (40\%, 35\% and 40\% by more than 1.8 points). At $n=128$ the harmful share is 34 to
39\% for all three. In whole runs on a reused selection set the same rules brought no gain
(Section~\ref{sec:v2}); the replay draws a fresh selection set for every decision, so it contains no lock-in, and its
candidates come from high-resolution trajectories. Table~\ref{tab:decisions} gives the full comparison. The gold set of
each oracle setting is split once into halves $G_1$ and $G_2$. Priors, tuned significance levels and tuned thresholds are
estimated on $G_1$ from other runs only (leave-one-run-out; across inference stacks for the 1.5B model; GSM8K has a
single oracle run and therefore no pilot). Each decision is made on $n$ items drawn from $G_2$ and credited with its held-out
effect on the remaining items of $G_2$; intervals are cluster bootstraps over generations. Since each setting has
only one or two oracle runs, whose generations share a trajectory, the intervals are conditional on those
trajectories and understate run-to-run variation. The nonparametric variant of the gate replaces the Gaussian prior
by the pilot's empirical effects but treats the candidates' measurement errors as independent, ignoring the
incumbent's shared error. The Gaussian priors in this replay subtract a fixed approximate variance
$0.15/|G_1|$ from the within-generation effect variance and floor the result at $10^{-4}$; this approximation
differs from the native-loop pilot fitting. The in-sample reference prior uses the evaluated runs' own held-out effects on $G_1$; in
settings with a single oracle run its phase-1 prior therefore contains the evaluated generation's own effects. The tuned $\alpha$ is 0.2
to 0.5 on TREC and 0.01 on Banking77. The tuned McNemar test improves on $\alpha=0.05$ by 0.07 to 0.21 points per
decision (significantly for $n\ge32$), and the Bayes gate by 0.09 to 0.20 (no interval excludes zero).

\begin{table}[htbp]
\centering
\caption{\textbf{Value of one acceptance decision} (held-out improvement in points per decision; mean over the seven oracle
settings, six for pilot-based rules since GSM8K has a single oracle run; differences over common settings;
95\% cluster-bootstrap intervals over generations, conditional on the oracle trajectories). $^\dagger$This row only: prior estimated in-sample from the evaluated runs, so the
rule is not deployable.}
\label{tab:decisions}
\small
\setlength{\tabcolsep}{3pt}
\resizebox{\textwidth}{!}{\begin{tabular}{lcccc}
\toprule
Rule & $n{=}16$ & $n{=}32$ & $n{=}64$ & $n{=}128$ \\
\midrule
Greedy keep-if-better & -0.47 {\scriptsize[-0.82, -0.03]} & -0.08 {\scriptsize[-0.46, +0.37]} & +0.24 {\scriptsize[-0.15, +0.71]} & +0.50 {\scriptsize[+0.11, +0.96]} \\
McNemar, $\alpha{=}.05$ & +0.06 {\scriptsize[+0.00, +0.14]} & +0.20 {\scriptsize[+0.04, +0.43]} & +0.34 {\scriptsize[+0.09, +0.67]} & +0.49 {\scriptsize[+0.18, +0.88]} \\
McNemar, $\alpha{=}.20$ & +0.12 {\scriptsize[-0.05, +0.34]} & +0.31 {\scriptsize[+0.04, +0.63]} & +0.43 {\scriptsize[+0.12, +0.82]} & +0.58 {\scriptsize[+0.25, +0.99]} \\
PACE e-process & +0.02 {\scriptsize[+0.00, +0.04]} & +0.09 {\scriptsize[+0.01, +0.23]} & +0.25 {\scriptsize[+0.04, +0.52]} & +0.43 {\scriptsize[+0.13, +0.79]} \\
Select-then-confirm & +0.06 {\scriptsize[-0.09, +0.26]} & +0.21 {\scriptsize[-0.03, +0.52]} & +0.35 {\scriptsize[+0.07, +0.71]} & +0.55 {\scriptsize[+0.21, +0.95]} \\
McNemar, pilot-tuned $\alpha$ & +0.19 {\scriptsize[-0.02, +0.43]} & +0.44 {\scriptsize[+0.12, +0.82]} & +0.56 {\scriptsize[+0.17, +1.02]} & +0.64 {\scriptsize[+0.25, +1.10]} \\
Threshold on $\hat\Delta$, pilot-tuned & +0.15 {\scriptsize[-0.07, +0.40]} & +0.35 {\scriptsize[-0.02, +0.78]} & +0.49 {\scriptsize[+0.09, +0.95]} & +0.60 {\scriptsize[+0.20, +1.03]} \\
Pilot-calibrated Bayes gate & +0.28 {\scriptsize[-0.03, +0.63]} & +0.40 {\scriptsize[+0.06, +0.82]} & +0.52 {\scriptsize[+0.14, +0.97]} & +0.66 {\scriptsize[+0.25, +1.12]} \\
\quad nonparametric prior & +0.20 {\scriptsize[-0.09, +0.52]} & +0.31 {\scriptsize[-0.02, +0.70]} & +0.51 {\scriptsize[+0.13, +0.95]} & +0.64 {\scriptsize[+0.25, +1.08]} \\
Bayes gate, in-sample prior$^\dagger$ & +0.21 {\scriptsize[-0.04, +0.52]} & +0.31 {\scriptsize[-0.01, +0.68]} & +0.47 {\scriptsize[+0.10, +0.90]} & +0.64 {\scriptsize[+0.25, +1.08]} \\
Best candidate (oracle) & +1.40 {\scriptsize[+1.00, +1.87]} & +1.42 {\scriptsize[+1.02, +1.89]} & +1.48 {\scriptsize[+1.08, +1.96]} & +1.61 {\scriptsize[+1.21, +2.07]} \\
\midrule
Pilot gate $-$ greedy & +0.77 {\scriptsize[+0.60, +0.93]} & +0.43 {\scriptsize[+0.30, +0.56]} & +0.21 {\scriptsize[+0.10, +0.30]} & +0.08 {\scriptsize[-0.00, +0.15]} \\
Tuned McNemar $-$ greedy & +0.68 {\scriptsize[+0.43, +0.91]} & +0.47 {\scriptsize[+0.31, +0.62]} & +0.25 {\scriptsize[+0.16, +0.34]} & +0.06 {\scriptsize[-0.04, +0.16]} \\
Pilot gate $-$ McNemar $.05$ & +0.20 {\scriptsize[-0.04, +0.48]} & +0.17 {\scriptsize[-0.03, +0.40]} & +0.12 {\scriptsize[-0.01, +0.27]} & +0.09 {\scriptsize[-0.01, +0.19]} \\
Tuned McNemar $-$ McNemar $.05$ & +0.11 {\scriptsize[-0.03, +0.28]} & +0.21 {\scriptsize[+0.06, +0.40]} & +0.16 {\scriptsize[+0.02, +0.32]} & +0.07 {\scriptsize[+0.01, +0.14]} \\
Pilot gate $-$ tuned McNemar & +0.09 {\scriptsize[-0.04, +0.24]} & -0.04 {\scriptsize[-0.13, +0.05]} & -0.04 {\scriptsize[-0.10, +0.01]} & +0.02 {\scriptsize[-0.05, +0.09]} \\
\bottomrule
\end{tabular}
}
\end{table}

\paragraph{Does a pilot transfer?}
A pilot on the same model and task with a large evaluation set is a strong requirement. Table~\ref{tab:transfer}
repeats the decision-level analysis above with the prior, the tuned $\alpha$ and the tuned
threshold taken from runs of the \emph{other} model on the same task (1.5B$\leftrightarrow$7B) or of the other task
with the same model (TREC$\leftrightarrow$Banking77; GSM8K from 1.5B TREC). Across models, the Bayes gate still
avoids greedy's losses at small $n$ and is at least as good as McNemar at $\alpha=0.05$ (significantly so at
$n=128$), while the tuned test transfers less well at $n=16$. Across tasks, only the protection against greedy's
losses transfers, and the gate is no better than McNemar at $\alpha=0.05$. A pilot is therefore useful when it
matches the task, and a conservative default is the safer choice when it does not.

\begin{table}[htbp]
\centering
\caption{\textbf{Prior transfer} (exploratory; value per decision in points, 95\% cluster-bootstrap intervals over
generations, conditional on the oracle trajectories).}
\label{tab:transfer}
\small
\setlength{\tabcolsep}{3pt}
\resizebox{\textwidth}{!}{\begin{tabular}{lcccc}
\toprule
Rule & $n{=}16$ & $n{=}32$ & $n{=}64$ & $n{=}128$ \\
\midrule
\multicolumn{5}{l}{\emph{Pilot from the other model (same task; 4 settings)}} \\
Greedy & -0.62 {\scriptsize[-1.08, -0.08]} & -0.10 {\scriptsize[-0.57, +0.47]} & +0.24 {\scriptsize[-0.26, +0.83]} & +0.48 {\scriptsize[+0.01, +1.06]} \\
McNemar $.05$ & +0.06 {\scriptsize[+0.01, +0.15]} & +0.16 {\scriptsize[+0.01, +0.34]} & +0.33 {\scriptsize[+0.03, +0.75]} & +0.51 {\scriptsize[+0.09, +1.06]} \\
McNemar, tuned $\alpha$ & -0.05 {\scriptsize[-0.19, +0.14]} & +0.28 {\scriptsize[-0.08, +0.74]} & +0.45 {\scriptsize[+0.01, +1.00]} & +0.56 {\scriptsize[+0.12, +1.13]} \\
Threshold, tuned & +0.00 {\scriptsize[-0.20, +0.28]} & +0.16 {\scriptsize[-0.15, +0.56]} & +0.46 {\scriptsize[+0.02, +1.03]} & +0.51 {\scriptsize[+0.07, +1.08]} \\
Pilot Bayes gate & +0.25 {\scriptsize[-0.08, +0.66]} & +0.39 {\scriptsize[-0.00, +0.90]} & +0.50 {\scriptsize[+0.05, +1.06]} & +0.62 {\scriptsize[+0.18, +1.19]} \\
Pilot gate $-$ greedy & +0.88 {\scriptsize[+0.63, +1.08]} & +0.49 {\scriptsize[+0.30, +0.67]} & +0.27 {\scriptsize[+0.14, +0.38]} & +0.14 {\scriptsize[+0.06, +0.22]} \\
Pilot gate $-$ McNemar $.05$ & +0.19 {\scriptsize[-0.09, +0.56]} & +0.23 {\scriptsize[-0.02, +0.57]} & +0.17 {\scriptsize[-0.00, +0.39]} & +0.11 {\scriptsize[+0.01, +0.22]} \\
\midrule
\multicolumn{5}{l}{\emph{Pilot from the other task (same model; 5 settings)}} \\
Greedy & -0.56 {\scriptsize[-0.97, -0.05]} & -0.16 {\scriptsize[-0.56, +0.36]} & +0.15 {\scriptsize[-0.28, +0.70]} & +0.39 {\scriptsize[-0.01, +0.94]} \\
McNemar $.05$ & +0.04 {\scriptsize[-0.00, +0.12]} & +0.13 {\scriptsize[+0.02, +0.29]} & +0.26 {\scriptsize[+0.03, +0.62]} & +0.41 {\scriptsize[+0.08, +0.88]} \\
McNemar, tuned $\alpha$ & -0.05 {\scriptsize[-0.18, +0.11]} & +0.02 {\scriptsize[-0.13, +0.21]} & +0.18 {\scriptsize[-0.03, +0.45]} & +0.46 {\scriptsize[+0.14, +0.92]} \\
Threshold, tuned & +0.04 {\scriptsize[-0.15, +0.31]} & +0.18 {\scriptsize[-0.09, +0.54]} & +0.35 {\scriptsize[+0.02, +0.80]} & +0.52 {\scriptsize[+0.18, +0.99]} \\
Pilot Bayes gate & +0.04 {\scriptsize[-0.16, +0.31]} & +0.22 {\scriptsize[-0.04, +0.59]} & +0.31 {\scriptsize[+0.01, +0.76]} & +0.43 {\scriptsize[+0.08, +0.92]} \\
Pilot gate $-$ greedy & +0.60 {\scriptsize[+0.26, +0.89]} & +0.38 {\scriptsize[+0.12, +0.63]} & +0.16 {\scriptsize[-0.07, +0.38]} & +0.04 {\scriptsize[-0.13, +0.21]} \\
Pilot gate $-$ McNemar $.05$ & -0.01 {\scriptsize[-0.18, +0.19]} & +0.09 {\scriptsize[-0.07, +0.30]} & +0.05 {\scriptsize[-0.06, +0.17]} & +0.03 {\scriptsize[-0.09, +0.15]} \\
\bottomrule
\end{tabular}
}
\end{table}

\paragraph{Whole-run results of all acceptance rules (exploratory).}
Tables~\ref{tab:runs_a}--\ref{tab:runs_b} report, for every rule and setting of the exploratory vLLM experiments, the
final held-out gain over the seed instruction (mean $\pm$ s.e. over 3 seeds), the final gap between the loop's
selection-set score and its held-out accuracy, and the number of harmful commits out of all commits (summed over
seeds). Whole-run differences between rules are mostly small relative to seed variation, and many gains arise in
the first few decisions, consistently with lock-in. Fixed-level tests rarely commit at $n=16$ and therefore stay near the seed,
which is optimal on Banking77 (where few proposals help) and costly on TREC; greedy's harmful commits and
self-reported gaps are largest at small $n$. The online EB gate did not improve on greedy in whole runs. On
1.5B-TREC at $n=64$ it ended 6.2 points below greedy ($+11.6\pm0.9$ vs.\ $+17.8\pm0.7$; lower in all three seeds,
paired $t$-test $p=0.016$), which we attribute to its
within-run estimate of $s^2$ from a handful of candidates, which is unreliable; we therefore evaluate only the
pilot-calibrated gate in the confirmatory study.

\begin{table}[htbp]
\centering
\caption{Whole runs, 1.5B, vLLM: held-out gain over the seed (points; mean $\pm$ s.e.\ over seeds), final
selection-set minus held-out gap (points), and harmful commits out of all commits, summed over seeds. RM runs use a
per-generation budget $B=256$ and are listed under $n=64$.}
\label{tab:runs_a}
\small
\begin{tabular}{llcccc}
\toprule
Setting & Rule & runs & held-out gain & proxy$-$held-out & harmful / commits \\
\midrule
1.5B TREC, $n{=}16$ & Greedy & 3 & +15.8$\pm$3.6 & 15.8 & 2 / 10 \\
 & McNemar & 3 & +11.3$\pm$5.8 & 7.7 & 0 / 2 \\
 & E-process & 3 & +0.0$\pm$0.0 & -1.8 & 0 / 0 \\
 & EB gate & 3 & +5.6$\pm$1.5 & 17.6 & 0 / 3 \\
\midrule
1.5B TREC, $n{=}64$ & Greedy & 3 & +17.8$\pm$0.7 & 7.4 & 2 / 13 \\
 & McNemar & 3 & +12.6$\pm$2.4 & 1.6 & 0 / 3 \\
 & E-process & 3 & +16.0$\pm$1.4 & 0.3 & 0 / 3 \\
 & EB gate & 3 & +11.6$\pm$0.9 & 4.5 & 1 / 4 \\
 & EB + RM & 3 & +16.2$\pm$1.9 & 5.3 & 0 / 6 \\
\midrule
1.5B Banking77, $n{=}16$ & Greedy & 3 & -5.4$\pm$2.9 & 10.3 & 5 / 5 \\
 & McNemar & 3 & +0.0$\pm$0.0 & -5.6 & 0 / 0 \\
 & E-process & 3 & +0.0$\pm$0.0 & -5.6 & 0 / 0 \\
 & EB gate & 3 & -0.5$\pm$0.5 & -3.0 & 1 / 1 \\
\midrule
1.5B Banking77, $n{=}64$ & Greedy & 3 & -3.6$\pm$2.5 & 1.1 & 5 / 5 \\
 & McNemar & 3 & +0.0$\pm$0.0 & -6.6 & 0 / 0 \\
 & E-process & 3 & +0.0$\pm$0.0 & -6.6 & 0 / 0 \\
 & EB gate & 3 & -1.9$\pm$1.2 & 0.5 & 2 / 2 \\
 & EB + RM & 3 & -0.6$\pm$0.3 & 0.4 & 2 / 3 \\
\midrule
1.5B GSM8K, $n{=}16$ & Greedy & 3 & -0.8$\pm$3.7 & 18.0 & 4 / 9 \\
 & EB gate & 3 & +2.4$\pm$1.0 & 14.9 & 2 / 5 \\
\midrule
1.5B GSM8K, $n{=}64$ & Greedy & 3 & +0.4$\pm$1.9 & 7.4 & 4 / 9 \\
 & McNemar & 3 & +4.3$\pm$1.6 & 6.8 & 1 / 4 \\
 & EB gate & 3 & +1.8$\pm$2.5 & 6.8 & 3 / 7 \\
 & EB + RM & 3 & -0.7$\pm$1.9 & 8.6 & 3 / 5 \\
\bottomrule
\end{tabular}

\end{table}

\begin{table}[htbp]
\centering
\caption{Whole runs, 7B, vLLM (same format as Table~\ref{tab:runs_a}).}
\label{tab:runs_b}
\small
\setlength{\tabcolsep}{3pt}
\begin{tabular}{llcccc}
\toprule
Setting & Rule & runs & held-out gain & proxy$-$held-out & harmful / commits \\
\midrule
7B TREC, $n{=}16$ & Greedy & 3 & +5.9$\pm$1.7 & 16.0 & 1 / 5 \\
 & McNemar & 3 & +1.0$\pm$1.0 & 10.7 & 0 / 1 \\
 & EB gate & 3 & +8.9$\pm$4.2 & 25.7 & 3 / 7 \\
\midrule
7B TREC, $n{=}64$ & Greedy & 3 & +15.4$\pm$3.2 & 10.9 & 4 / 14 \\
 & McNemar & 3 & +8.9$\pm$3.0 & 6.9 & 0 / 3 \\
 & E-process & 3 & +5.9$\pm$5.9 & 2.6 & 0 / 1 \\
 & EB gate & 3 & +13.9$\pm$4.2 & 10.2 & 3 / 10 \\
 & EB + RM & 3 & +14.1$\pm$2.8 & 4.7 & 2 / 10 \\
\midrule
7B Banking77, $n{=}16$ & Greedy & 3 & -0.2$\pm$1.3 & 3.1 & 1 / 3 \\
 & McNemar & 3 & +0.0$\pm$0.0 & -3.4 & 0 / 0 \\
 & EB gate & 3 & +0.0$\pm$0.6 & 2.8 & 1 / 2 \\
\midrule
7B Banking77, $n{=}64$ & Greedy & 3 & +0.8$\pm$0.3 & -0.5 & 2 / 6 \\
 & McNemar & 3 & +0.5$\pm$0.5 & -1.3 & 0 / 1 \\
 & E-process & 3 & +0.0$\pm$0.0 & -3.9 & 0 / 0 \\
 & EB gate & 3 & +1.2$\pm$0.4 & -2.0 & 0 / 3 \\
 & EB + RM & 3 & +0.9$\pm$0.2 & -4.7 & 0 / 3 \\
\bottomrule
\end{tabular}

\end{table}

\paragraph{Fresh versus reused selection sets.}
Re-drawing the selection set every generation (``greedy, fresh $D$'', llama.cpp replication, TREC $n=64$) avoids
carrying the incumbent's earlier selection error into the next comparison. The loop commits twice as often
(33 vs.\ 16 commits over three seeds), 39\% of its commits are harmful
(vs.\ 25\% with reuse; 30\% vs.\ 12.5\% at the $-1.8$-point threshold), and its final held-out gain is lower ($+14.4\pm1.1$ vs.\ $+21.3\pm1.4$ points),
consistent with Proposition~\ref{prop:goodhart} when most proposals are harmful and each comparison is fresh noise.
This is one exploratory setting with three seeds and the feedback confound of the exploratory design; the difference
in harmful shares is not significant (Fisher $p=0.36$), and the fresh-set runs had fewer valid candidates per
generation (2.3 against 3.1).

\paragraph{Development-seed failure of the unsafeguarded EB gate.}
On the development seed (TREC, llama.cpp), the first version of the online EB gate, combined with \emph{resolution
matching} (increasing $n$ across generations to hold the estimated $h_w^2$ fixed), peaked at $+18.4$ and ended at
$+5.8$ points with 3 harmful commits out of 8: after two generations its estimate of $s^2$ collapsed to zero, the
posterior reduced to a stale positive $\hat\mu$, and the gate committed candidates that \emph{measured} as
regressions. We added the safeguard of Section~\ref{sec:method} (never commit a measured regression) and capped the
growth of $n$, and ran the safeguarded gate on all seeds and tasks; this change was post hoc, and the EB-gate
entries of the exploratory whole-run tables average over all seeds, including the development seed. The
unsafeguarded runs are listed in Table~\ref{tab:runs_l} as an ablation.

\paragraph{Allocation sweeps.}
Table~\ref{tab:alloc} splits a fixed per-generation budget of $B=256$ evaluations into $K$ candidates of $n$ items
(online EB gate). Differences are within seed variation; for 7B the extremes ($K=1$ and $K=16$) do worst, consistent
with an interior optimum, but the sweep is too small to be conclusive. An earlier sweep on the llama.cpp stack (1.5B
TREC, two seeds, $T=25$) gave held-out gains of 16.7, 13.9, 24.2 and 14.6 points for $(K,n)=(1,256)$, $(2,128)$,
$(8,32)$ and $(16,16)$, again within seed variation.

\begin{table}[htbp]
\centering
\caption{Held-out gain (pts, mean $\pm$ s.e., number of seeds) for splits of $B=256$ evaluations per generation.}
\label{tab:alloc}
\small
\setlength{\tabcolsep}{2pt}
\begin{tabular}{lccccc}
\toprule
Model & $K{\times}n=16{\times}16$ & $8{\times}32$ & $4{\times}64$ & $2{\times}128$ & $1{\times}256$ \\
\midrule
1.5B TREC & +16.3{\scriptsize$\pm$1.9} (3) & +13.6{\scriptsize$\pm$2.0} (3) & +11.6{\scriptsize$\pm$0.9} (3) & +14.8{\scriptsize$\pm$1.3} (3) & +13.7{\scriptsize$\pm$2.2} (3) \\
7B TREC & +8.8{\scriptsize$\pm$4.2} (2) & +11.5{\scriptsize$\pm$7.3} (2) & +13.9{\scriptsize$\pm$4.2} (3) & +12.6{\scriptsize$\pm$5.3} (2) & +6.6{\scriptsize$\pm$3.8} (2) \\
\bottomrule
\end{tabular}

\end{table}

\section{Replication with a second inference stack}
\label{app:repl}

We repeated the 1.5B classification experiments with an independent inference stack: 8-bit GGUF weights served by
llama.cpp on a consumer GPU (RTX 3080 Ti), $T=25$ generations, otherwise identical code. Two oracle settings are
included in Figure~\ref{fig:breeder} (``llama.cpp''); the distribution-aware prediction has mean absolute error
0.014 (TREC) and 0.048 (Banking77). Greedy loops again gain more held-out accuracy with larger $n$ on TREC
($+9.1$ at $n=16$ vs.\ $+25.7$ at $n=256$) while their self-reported gap shrinks (14.4 vs.\ 5.4 points), and on
Banking77 they again lose held-out accuracy ($-3.8$ at $n=16$ and $n=64$, $-1.4$ at $n=256$). Whole-run results
of the acceptance rules are in Table~\ref{tab:runs_l}; on this stack the safeguarded EB gate matches greedy at
$n=64$ and is within seed variation of it at $n=16$ on TREC ($+10.8\pm4.0$ vs.\ $+9.1\pm2.5$), and EB with
resolution matching attains the highest TREC gain among rules with 256 evaluations per generation
($+24.8\pm0.3$; greedy at $n=256$, with 1024 per generation, reaches $+25.7$).

\begin{table}[htbp]
\centering
\caption{Whole runs, 1.5B, llama.cpp replication ($T=25$; same format as Table~\ref{tab:runs_a}).}
\label{tab:runs_l}
\small
\setlength{\tabcolsep}{3pt}
\begin{tabular}{llcccc}
\toprule
Setting & Rule & runs & held-out gain & proxy$-$held-out & harmful / commits \\
\midrule
1.5B TREC, $n{=}16$ & Greedy & 3 & +9.1$\pm$2.5 & 14.4 & 1 / 9 \\
 & McNemar & 3 & +0.0$\pm$0.0 & -3.6 & 0 / 0 \\
 & E-process & 3 & +0.0$\pm$0.0 & -3.6 & 0 / 0 \\
 & EB gate & 3 & +10.8$\pm$4.0 & 4.3 & 0 / 5 \\
 & EB, no safeguard & 2 & +12.8$\pm$7.4 & 8.7 & 0 / 5 \\
\midrule
1.5B TREC, $n{=}64$ & Greedy & 3 & +21.3$\pm$1.4 & 6.9 & 4 / 16 \\
 & Greedy, fresh $D$ & 3 & +14.4$\pm$1.1 & 6.0 & 13 / 33 \\
 & McNemar & 3 & +17.6$\pm$2.8 & 6.5 & 2 / 6 \\
 & E-process & 3 & +18.8$\pm$3.0 & 7.8 & 0 / 4 \\
 & EB gate & 3 & +21.2$\pm$1.2 & 7.0 & 3 / 15 \\
 & EB + RM & 3 & +24.8$\pm$0.3 & 4.1 & 2 / 11 \\
 & EB, no safeguard & 1 & +21.7$\pm$0.0 & -0.3 & 0 / 6 \\
 & EB + RM, no safeguard & 1 & +5.8$\pm$0.0 & 5.8 & 3 / 8 \\
\midrule
1.5B Banking77, $n{=}16$ & Greedy & 3 & -3.8$\pm$3.5 & 12.0 & 3 / 4 \\
 & McNemar & 3 & +0.0$\pm$0.0 & -2.3 & 0 / 0 \\
 & E-process & 3 & +0.0$\pm$0.0 & -2.3 & 0 / 0 \\
 & EB gate & 3 & -2.6$\pm$2.6 & 6.6 & 2 / 2 \\
\midrule
1.5B Banking77, $n{=}64$ & Greedy & 3 & -3.8$\pm$2.7 & 4.1 & 5 / 8 \\
 & McNemar & 3 & +0.0$\pm$0.0 & -6.9 & 0 / 0 \\
 & E-process & 3 & +0.0$\pm$0.0 & -6.9 & 0 / 0 \\
 & EB gate & 3 & -3.1$\pm$2.4 & 2.4 & 3 / 5 \\
 & EB + RM & 3 & -1.7$\pm$0.8 & -3.8 & 3 / 4 \\
\bottomrule
\end{tabular}

\end{table}

\end{document}